\documentclass[10pt,journal,compsoc]{IEEEtran}

\ifCLASSOPTIONcompsoc
  \usepackage[nocompress]{cite}
\else
  \usepackage{cite}
\fi

\ifCLASSINFOpdf
   \usepackage[pdftex]{graphicx}
\else
\fi

\usepackage{csquotes}
\usepackage{amsmath}
\usepackage{wrapfig}
\usepackage{cases}
\usepackage[ruled,linesnumbered]{algorithm2e}
\usepackage{multirow}
\usepackage{booktabs}
\usepackage{subcaption}
\usepackage{color}

\begin{document}

%
\title{MLink: Linking Black-Box Models from Multiple Domains for Collaborative Inference~\thanks{This article is a substantially extended and revised version of Yuan et al.~\cite{mlink}, which appeared in the proceedings of the 36th AAAI Conference on Artificial Intelligence (AAAI '22).}}
%
%
%
%

\author{Mu~Yuan,~%
        Lan~Zhang,~\IEEEmembership{Member,~IEEE,}
        Zimu~Zheng,~\IEEEmembership{Member,~IEEE,}
        Yi-Nan~Zhang,
        and~Xiang-Yang~Li,~\IEEEmembership{Fellow,~IEEE}%
\IEEEcompsocitemizethanks{
\IEEEcompsocthanksitem Lan Zhang and Xiang-Yang Li are the corresponding authors.
\IEEEcompsocthanksitem Lan Zhang is with the School of Computer Science and Technology, University of Science and Technology of China, Hefei, China, and Institute of Artificial Intelligence, Hefei Comprehensive National Science Center, Hefei, China.\protect\\
E-mail: zhanglan@ustc.edu.cn
\IEEEcompsocthanksitem Mu Yuan, Yi-Nan Zhang, and Xiang-Yang Li are
with the School of Computer Science and Technology, University of Science and Technology of China, Hefei, China.\protect\\
E-mail: ym0813@mail.ustc.edu.cn, zhangyinan@mail.ustc.edu.cn, xiangyangli@ustc.edu.cn 
\IEEEcompsocthanksitem Zimu Zheng is with the Edge Cloud Innovation Lab, Huawei Cloud.\protect\\
Email: zimu.zheng@huawei.com
}
\thanks{
}
}

\IEEEtitleabstractindextext{%
\begin{abstract}
The cost efficiency of model inference is critical to real-world machine learning (ML) applications, especially for delay-sensitive tasks and resource-limited devices.
A typical dilemma is: in order to provide complex intelligent services (e.g. smart city), we need inference results of multiple ML models, but the cost budget (e.g. GPU memory) is not enough to run all of them.
In this work, we study underlying relationships among black-box ML models and propose a novel learning task: model linking, which aims to bridge the knowledge of different black-box models by learning mappings (dubbed model links) between their output spaces.
We propose the design of model links which supports linking heterogeneous black-box ML models.
Also, in order to address the distribution discrepancy challenge, we present adaptation and aggregation methods of model links.
Based on our proposed model links, we developed a scheduling algorithm, named \textit{MLink}.
Through collaborative multi-model inference enabled by model links, \textit{MLink} can improve the accuracy of obtained inference results under the cost budget.
We evaluated \textit{MLink} on a multi-modal dataset with seven different ML models and two real-world video analytics systems with six ML models and 3,264 hours of video.
Experimental results show that our proposed model links can be effectively built among various black-box models.
Under the budget of GPU memory, \textit{MLink} can save 66.7\% inference computations while preserving 94\% inference accuracy, which outperforms multi-task learning, deep reinforcement learning-based scheduler and frame filtering baselines.
\end{abstract}

}

\maketitle

\IEEEdisplaynontitleabstractindextext

%
\IEEEpeerreviewmaketitle

\IEEEraisesectionheading{\section{Introduction}\label{sec:introduction}}

\IEEEPARstart{M}{ulti-model} inference workloads are increasingly prevalent, e.g., smart speaker assistants~\cite{smart-speaker}, smart cities~\cite{smartcity}, drone-based video monitoring~\cite{drone-video}, multi-modal autonomous driving~\cite{multi-modal-auto-drive}, etc.
Besides the accuracy of the trained models, costs in the inference phase can become the bottleneck to the quality of services, especially for delay-sensitive tasks and resource-limited devices.

Towards cost-efficient inference, existing work explored various perspectives to achieve the resource-performance trade-offs.
Multi-task learning and zipping~\cite{multitask-zip,hierarchical-mtl,mtl-survey,mtl} can reduce the computing overheads by sharing neurons among different tasks;
Model compression~\cite{kd,resource-adadeep, adversarial-distillation,few-shot-compress} techniques attempt to eliminate parameters and connections not related to the inference accuracy; 
Inference reusing~\cite{reuse-foggycache,reuse-deep} approaches aim to avoid the same or similar computations;
Source filtering~\cite{reducto} methods try to transmit only necessary input data to backend ML models.
Adaptive configuration~\cite{adaconf} and multi-model scheduling~\cite{adams} were proposed to make inference workloads adaptive to the dynamics of input content.
We summarize them as answers to an interesting question:
\begin{displayquote}
   \textit{How to obtain as accurate inference results as possible without the exact execution of ML models?}
\end{displayquote}
From this perspective, multi-task learning and model compression generates a lighter model for the same inference task(s) by pruning the original model(s).
Inference reusing and source filtering techniques reuse previous inference results as the predicted results through analyzing the correlation between inputs. 
Based on the observation that, for some input data, the accuracy of expensive and cheap models is similar, adaptive configuration analyzes the input dynamics and predicts the inference results of expensive models by executing cheap ones.
Adaptive multi-model scheduling predicts unnecessary inference results as empty using the executed models' outputs as the hint information.

\begin{figure}[t]
    \centering
    \begin{subfigure}[b]{0.98\linewidth}
         \centering
         \includegraphics[width=\linewidth]{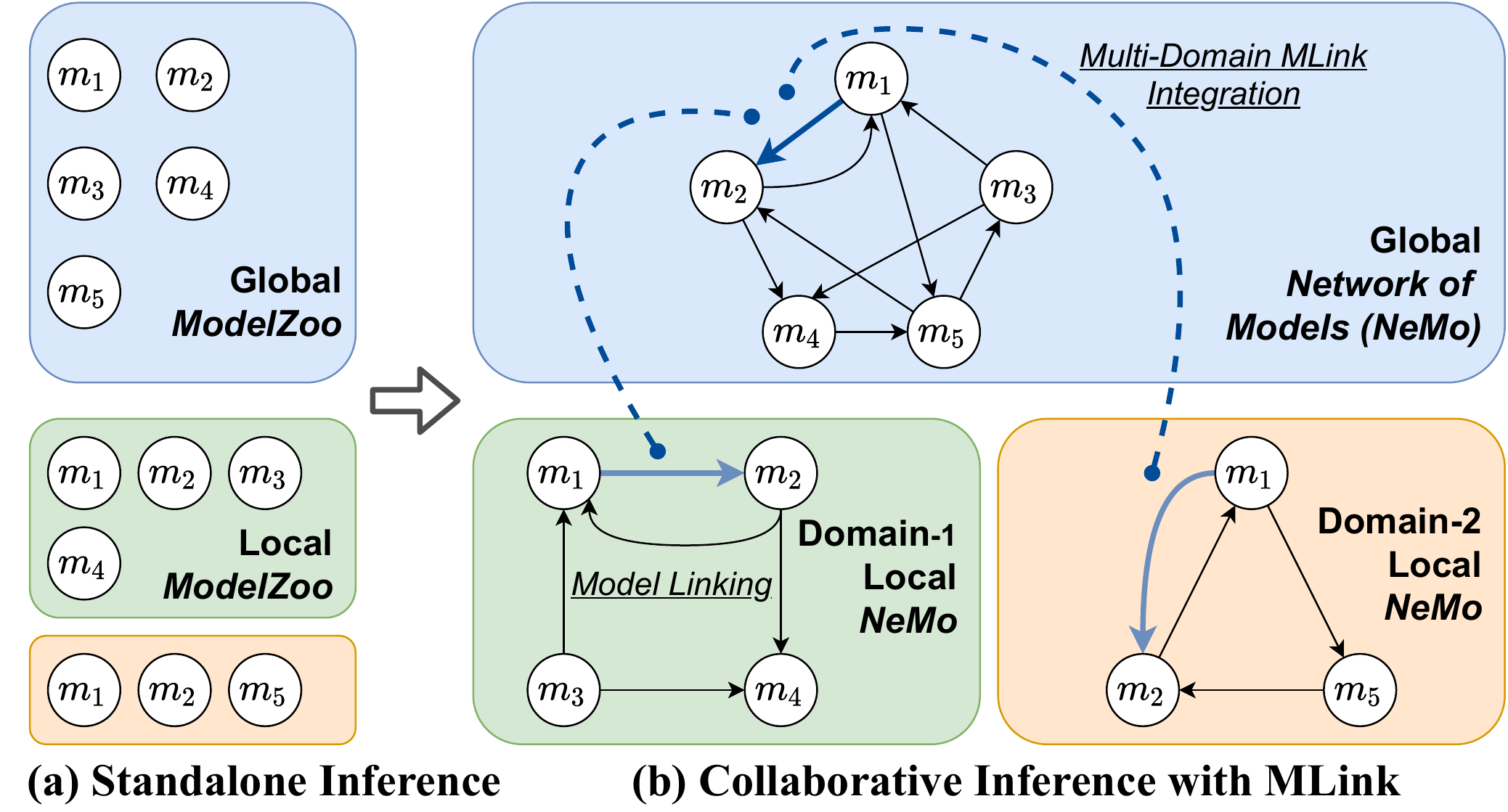}
     \end{subfigure}
\caption{From standalone inference to multi-domain collaborative inference based on \textit{MLink}. $m_1$-$m_5$ denotes ML models. Arrows refer to our proposed model links and blue dotted lines refer to the aggregation process of cross-domain model links.}
\label{fig:intro}
\end{figure}

We address this problem from a novel perspective: \emph{linking black-box models}.
We were motivated by the insight that even ML models that are different in input modalities, learning tasks, architectures, etc., can share knowledge with each other, since ML models are prone to ``overlearning''~\cite{Song2020Overlearning} and outputs of different models have semantic correlations~\cite{tl-survey}.
If we can effectively bridge the knowledge among ML models, we can directly predict inference results of remaining models based on executed models' outputs.
If the cost of this prediction is low, it is promising to improve the resulting accuracy of inference results from all models under a limited cost budget, compared to the original workflow of standalone inference where results of unexecuted models cannot be obtained at all.
Fig.~\ref{fig:intro} illustrates the transition from standalone inference to collaborative inference based on our proposed model linking approach.
To realize this vision, the following three main challenges need to be solved:

(1) \textbf{How to build knowledge-level links among black-box and highly different ML models?}
In practice, deployed ML models could have different architectures and input modalities, and they could be developed by different programming languages and ML frameworks.
The heterogeneity makes it challenging to design a general model of knowledge-level connections among ML models.
On the other hand, model linking should be non-intrusive to the original inference system and require as little model information and code modification as possible.
The black-box access of ML models bringing additional challenges to the design and implementation.

(2) \textbf{How to adapt model links to dynamic data distribution on the fly?}
Due to the dynamics of inputs, model linking faces an adaptation challenge similar to classical ML, namely domain adaptation~\cite{wang2018deep}.
Domain shift is typically caused by two factors, i.e., online dynamics of streaming content and differences in application scenarios.
For example, running a computer vision model on video streams captured from different cameras or at different times faces distribution shifts~\cite{online-kd}.

(3) \textbf{How to efficiently select models to be executed and models to be predicted?}
Given a set of ML models, after constructing model links among them, we need to select a proper subset of models to be executed under a certain cost budget, e.g. the allocatable GPU memory.
Highly efficient model selection is critical to the cost-performance trade-off, which is non-trivial due to its theoretical hardness. 

In this paper, we first formalize the model linking task and propose the design of model links which supports linking heterogeneous black-box ML models.
Then we present adaptation and aggregation methods of model links, covering both the online dynamics and cross-domain distribution shifts.
And we develop a model link-based algorithm, named \textit{MLink}, to schedule multi-model inference under a cost budget.
We evaluated our designs on a multi-modality dataset with seven different ML models, covering five classes of learning tasks and three types of input modalities.
Results show that our proposed model links can be effectively built among heterogeneous black-box models.
We evaluated \textit{MLink} on two real-world video analytics systems, one for the smart building and the other for city traffic monitoring, including six visual models and 3,264 hours of video from 58 cameras.
Experimental results show that our online adaptive training methods effectively improves the performance than the vanilla offline training.
And our aggregation approach achieves 7.9\% higher average accuracy than the original model.
Under the budget of GPU memory, \textit{MLink} outperforms baselines (multi-task learning~\cite{mtl-survey}, deep reinforcement learning-based scheduler~\cite{adams} and frame filtering~\cite{reducto}) and can save 66.7\% inference computation while preserving 94\% output accuracy.

\section{Problem Statement}
In this section, we define the model linking task and the inference under budgets problem.

\textbf{Model linking.}
Given a set of black-box ML models $F=\{f_i\}_{i=1}^k$, where $f_i:X_i\rightarrow Y_i$ is a function mapping the input to its inference result.
ML models can be highly heterogeneous, i.e., different input modalities, learning tasks, architectures, etc.
We only assume that input spaces $\{X_i\}_{i=1}^k$ are the same or aligned.
The case that different models share the same input spaces is common, e.g., multi-task learning-based robotics~\cite{mtl-survey,mtl} and multimedia advertising~\cite{video-ads}.
The aligned input spaces typically exists in the context of multi-modal scenarios, e.g., multi-modality event detection~\cite{zero-shot-event} and visual speech synthesis~\cite{multimodal-survey}.
In practice, synchronization in time can easily align inputs for many applications.
Moreover, approaches such as spatial alignment of multi-view videos~\cite{black2002multi} and audio-visual semantic alignment~\cite{wang2020alignnet} can be adopted for specific scenarios.
We define model linking as a function $g_{ij}: Y_i \rightarrow Y_j$, i.e., a mapping from the source model $f_i$'s output space to the target model $f_j$'s.
Then the composite function $g_{ij}\circ f_{i}:X_i\rightarrow Y_j$ can perform the inference computation of $f_j$.
Correspondingly, $g_{ji}$ links the knowledge of $f_i$ into $g_{ji}\circ f_{j}$.

\textbf{Multi-source model links ensemble.}
When the number of models $k\geq 3$, for one target model $f_j$, there could be multiple model links from different sources.
Let $A \subseteq F$ denote the set of source models.
Then for all $f_i \in A$, $g_{ij}\circ f_i$ performs the prediction task to $f_j$'s inference outputs.
The question that follows is, how do we determine the final prediction?
From the ensemble learning perspective, $\{g_{ij}\circ f_i\}_{f_i \in A}$ constitute a multi-expert model~\cite{moe}, which has the potential to perform better prediction with the multi-task \& multi-modal representation~\cite{multimodal-survey, mtl}.
We define $h_{A, j}$ as the ensemble model link from $A$ to $f_j$.
So the input of $h_{A, j}$ is the set of predictions by $g_{ij}$ where $f_i \in A$.
Note that if $A$ has only one element $f_i$, then $h_{A,j}=g_{ij}\circ f_{i}$.

\textbf{Multi-model inference under budget.}
The model links can be utilized to achieve resource-performance trade-offs of multi-model inference workloads.
Let $c(\cdot)$ denote the cost of running a function, e.g., GPU memory or inference time.
For resource-limited devices (e.g., smartwatches and mobile phones) and delay-sensitive tasks (e.g., real-time video analytics and audio assistant), there are certain constraints on the total cost.
We define $B$ as the cost budget and aim to maximize the inference accuracy under that budget.
Let $p(h_{A, j})$ denote the performance measure of the ensemble model link, which depends on the target model's task.
We assume the range of $p$ is normalized into $[0,1]$.
For example, the performance measure can be accuracy for classification task and bounding box IoU for the detection task.
Following previous efforts for optimizing the inference efficiency~\cite{adams, reducto}, the performance measure the consistency between obtained results and exact inference outputs, instead of ground-truth labels.
The multi-model inference under cost budget problem is formalized as:
\begin{equation}
\label{eq:opt}
\begin{split}
    \max\limits_{A\subseteq F}\overbrace{(\frac{1}{|F|}(\underbrace{\sum_{f_i \in A}1}_{\text{activated}} + \underbrace{\sum_{f_j\in F\setminus A} p(h_{A, j})}_{\text{predicted}} ))}^{\text{average output accuracy}}
    \\s.t.~~\underbrace{\sum_{f_i \in A} c(f_i)}_{\text{exact inference}} + \underbrace{\sum_{f_j \in F\setminus A} c(h_{A,j})}_{\text{model links}} \leq B.
\end{split}
\end{equation}
Under the cost budget, the optimization problem aims to maximize the average performance of all models $F$ by selecting an \textit{activated} subset $A$ to be executed.
For ease of description, we define the objective function as the \textit{output accuracy}.
Activated models do exact inference, so their performance scores are all 1.
Models that are not activated only participate in constructing model links and will not be executed during the inference phase; 
instead, they are predicted by the activated models via ensemble model links.
The cost of activated models is performing exact inference, while the cost of predicted models is from running model links.
So the model links should be both accurate and lightweight to reduce the cost while preserving the quality of the multi-model inference workloads.

\section{Black-Box Model Linking}
\label{sec:modellink}

In this section, we discuss the motivation of linking black-box models and present the theoretical analysis, architecture design, ensemble and training methods of model links.

\subsection{Motivational Study}
\begin{figure}[t]
    \centering
    \begin{subfigure}[b]{0.49\linewidth}
         \centering
         \includegraphics[width=\linewidth]{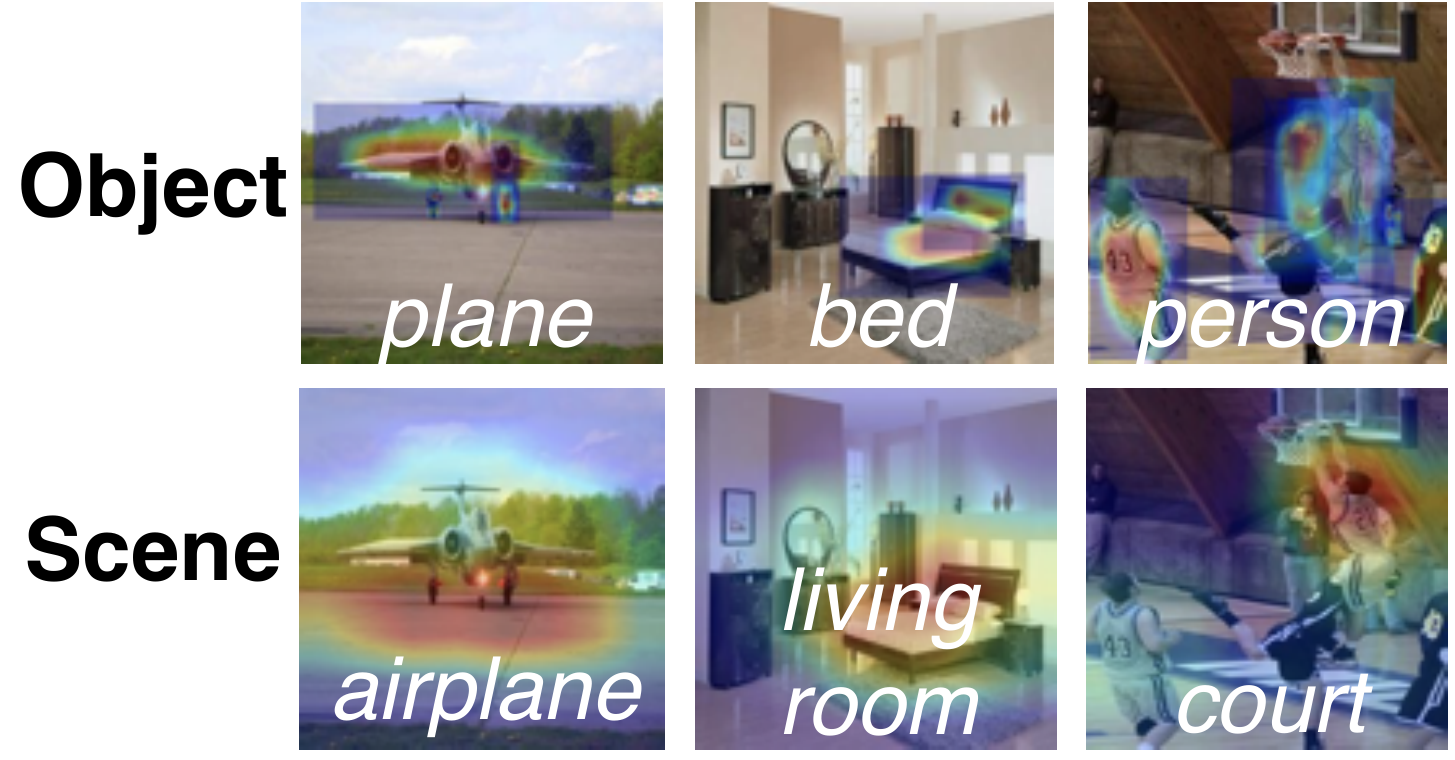}
         \caption{Attention heatmaps of Object and Scene models.}
         \label{fig:attention}
     \end{subfigure}
     \hfill
     \begin{subfigure}[b]{0.49\linewidth}
         \centering
         \includegraphics[width=\linewidth]{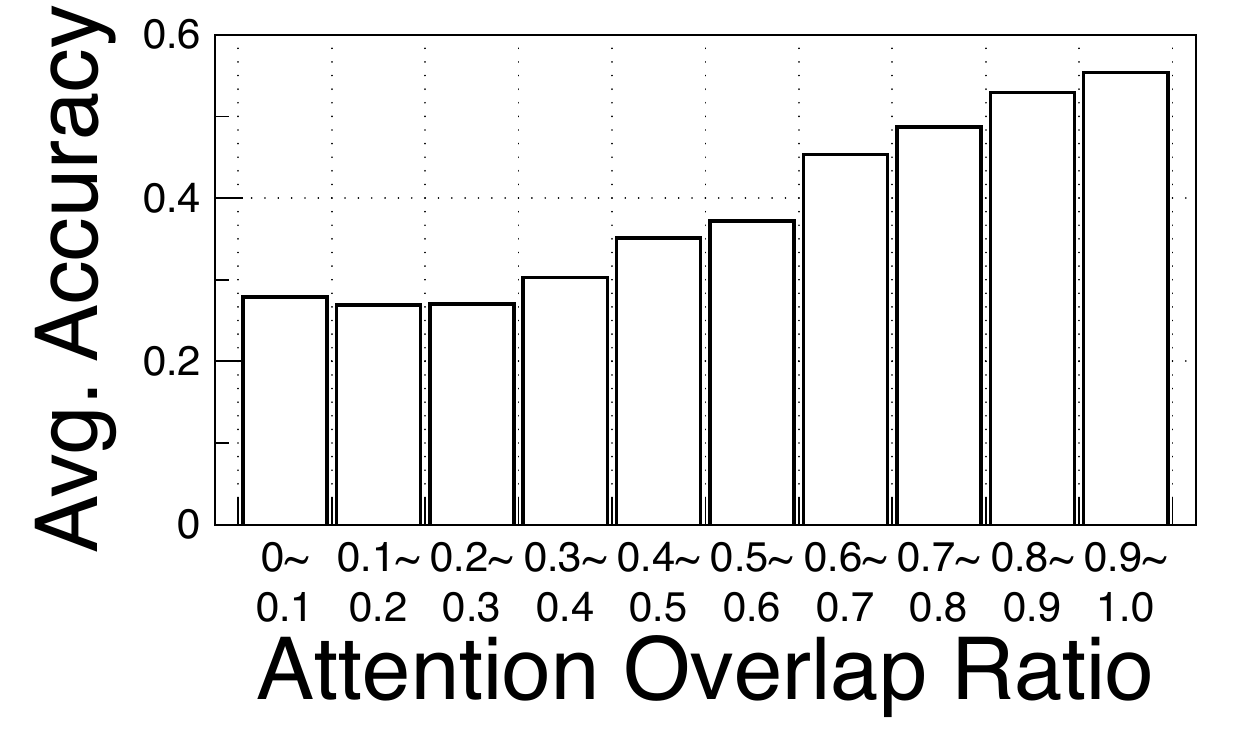}
         \caption{Scene-to-Object MLink accuracy vs. attention overlaps.}
         \label{fig:acc-ratio}
     \end{subfigure}
\caption{Inter-model Semantic Correlation}
\label{fig:motivation}
\end{figure}

When training ML models for different tasks, the ideal representation learned by them should be independent and disentangled~\cite{hjelm2018learning}, i.e. each model only learns the semantics that just covers its objective task.
However, due to the mismatched complexity of the data and the model, the machine learning process is prone to ``overlearning''~\cite{Song2020Overlearning}, which means that unintended semantics is encoded in the learned representation.
Besides, there exist semantic correlations among outputs of different tasks and different models may pay attention to the same content, e.g., the same regions in images. 
For example, in Fig.~\ref{fig:attention}, based on G-CAM~\cite{gcam}, we plot the attention heatmaps of YOLO-V3~\cite{yolov3} object detector and ResNet50~\cite{resnet} scene classifier on the same images, and their attention areas have much overlap.
We experimented on the correlation between the overlap ratio of attention heatmaps and the performance of model linking. 
For example, from the scene classification model to the object detection model, as shown in Fig.~\ref{fig:acc-ratio}, the accuracy of model links is obviously relevant with the overlap ratio ($(Map_{source} \land Map_{target} )/Map_{target} $). 
To a certain extent, it shows that the correlation learned by model links is similar with the semantic attention.
The ``overlearning'' characteristic and underlying semantic correlations among outputs make mappings from the same or aligned input space to different output spaces transferable~\cite{tl-survey}.

\subsection{Black-Box Output vs. Intermediate Representation}
A key design principle is that we only use the black-box output of the source model to for model linking.
Existing work has shown that by fine-tuning the last few layers~\cite{finetune}, the intermediate representation can be used to predict other different tasks.
However, in real applications, we often have to deal with the deployed models, which only provide a black-box inference API.
Compared with intermediate representation, the downstream black-box outputs do have weaker representation capability for general learning tasks.
But recent work~\cite{adams} shows that, given the same (or aligned) inputs, the executed models' outputs are very effective hints for scheduling unexecuted models.
The insight is that the correlation of black-box outputs between multiple tasks with the same input is more explicit and even stronger than the intermediate features.
And our experimental results also show that, using the same amount of training data, black-box model linking achieves higher accuracy than a knowledge distillation approach (see Fig.~\ref{fig:hw2-mlink}) and a multi-task learning approach (see Tab.~\ref{tab:schedule}).
Considering the better practicality and satisfactory accuracy, we select black-box outputs rather than intermediate representations for linking ML models.

\subsection{Sample Complexity Analysis}
Let $f\in \mathcal{F}$ denote task-specific parameters and $h$ denote shared parameters across tasks.
It has been proved that when the training data for $h$ is abundant, to achieve bounded prediction error on a new task only requires $C(\mathcal{F})$ sample complexity~\cite{tripuraneni2020theory}, where $C(\cdot)$ is the complexity of a hypothesis family.
Learning a model link $g_{ij} \in \mathcal{G}$ from source model $f_{i}\in \mathcal{F}_i$ to the target $f_j\in \mathcal{F}_j$ constitutes a compound learning model $g_{ij}\circ f_{i}$.
A lightweight design of model links can makes $C(\mathcal{G}) < C(\mathcal{F}_j)$ hold.
Therefore, applying the above result, model linking can significantly reduce the sample complexity to $C(\mathcal{G})$, compared with the $C(\mathcal{F}_j)$ complexity of learning the target model from scratch.
This result is also confirmed by our experiments: effective model links can be learnt by a very small amount (e.g. 1\%) of training samples (see Fig.\ref{fig:hw2-mlink}).

\subsection{Model Link Architecture}
Model links map between black-box models' output spaces, so the output format determines the architecture.
We classify output formats as the fixed-length vector and the variable-length sequence.
These two types of outputs could cover most ML models.
We propose four types of model link architectures based on best practices of similar learning tasks.

\textbf{Vec-to-Vec:} The model link maps from a vector-output source to a vector-output target.
We use a ReLU-activated multilayer perception (MLP) for the vec-to-vec model link.

\textbf{Seq-to-Vec:} The model link maps from a sequence-output source to a vector-output target.
We first use an embedding layer, which performs a matrix multiplication to transform the sequence into a fixed-size embedding.
Then we use an LSTM~\cite{lstm} layer followed by an MLP to generate the vector output.

\textbf{Vec-to-Seq:} The model link maps from a vector-output source to a sequence-output target.
We adopt the encoder-decoder framework, where an MLP serves as the encoder and the decoder consists of an embedding layer, an LSTM layer, an attention layer~\cite{attention}, and a fully-connected layer, in the forward order.

\textbf{Seq-to-Seq:} The model link maps from a sequence-output source to a sequence-output target.
We adopt the sequence-to-sequence framework~\cite{seq2seq}, where an embedding layer followed by an LSTM layer serves as the encoder and the decoder is the same as the one in the vec-to-seq model link.

The output activation functions are determined by the learning task of the target model. 
Softmax is used for single-label classification, and sigmoid is used for multi-label classification and sequence prediction. 
Linear activation works with regression and localization tasks.
In our implementation, the default number of hidden units is twice the length of the output dimension, which empirically achieved a good trade-off between effectiveness and efficiency.

\subsection{Ensemble of Multi-Source Links}
\label{subsec:ensemble}
The ensemble of multi-source model links has the potential to improve the prediction performance~\cite{multi-model-ensemble}, since cross-task and cross-modal representation capabilities could be beneficial.
For the target model $f_j$, given the set of sources $A$, we multiply outputs of $g_{ij}$ by trainable weights, where $f_i \in A$.
The weighted prediction is then activated according to $f_j$'s learning task.
The learned weights of $h_{A,j}$ can be used to ensemble model links from any subset of sources, i.e., $h_{A',j}, A'\subset A$.

\subsection{Training}
Classic knowledge distillation~\cite{kd} suggests that soft-label supervisions are better for training the ``student'' model, since the ``teacher'' model's outputs augment the hard-label space with relations among different classes.
Our experimental results show that this empirical experience still holds in the proposed model linking setting.
To train model links and the ensemble model, we collect $n$ inference results $\{\{y_i^j\}_{j=1}^k\}_{i=1}^n$ from $k$ models on the same or aligned inputs.
Given $f_i, f_j$ as the source and the target, respectively, the objective of training the model link $g_{ij}$ is:
\begin{equation}
\label{eq:loss1}
    \min \sum_{l=1}^n \mathcal{L}_j(g_{ij}(y_i^l), y_j^l),
\end{equation}
where the loss function $\mathcal{L}_j$ depends the learning task of the target model $f_j$.
Given $A, f_j$ as the set of sources and the target, respectively, the objective of training the ensemble model $h_{A,j}$ is:
\begin{equation}
\label{eq:loss2}
    \min \sum_{i=l}^n \mathcal{L}_j(h_{A,j}(\{y_i^l\}_{f_i \in A}), y_j^l).
\end{equation}
Both model links and ensemble models are optimized via gradient descent.
Note that if $A$ has only one element $f_i$, then the ensemble simply fits as an identity layer and $h_{A,j}=g_{ij}\circ f_i$.
\section{Model Link Adaptation and Aggregation}

In this section, we present designs for online adaptive training of model links.
And we discuss how to leverage model links for domain adaptation and propose an approach to aggregate cross-domain model links.

\begin{figure}[t]
    \centering
    \begin{subfigure}[b]{0.49\linewidth}
         \centering
         \includegraphics[width=\linewidth]{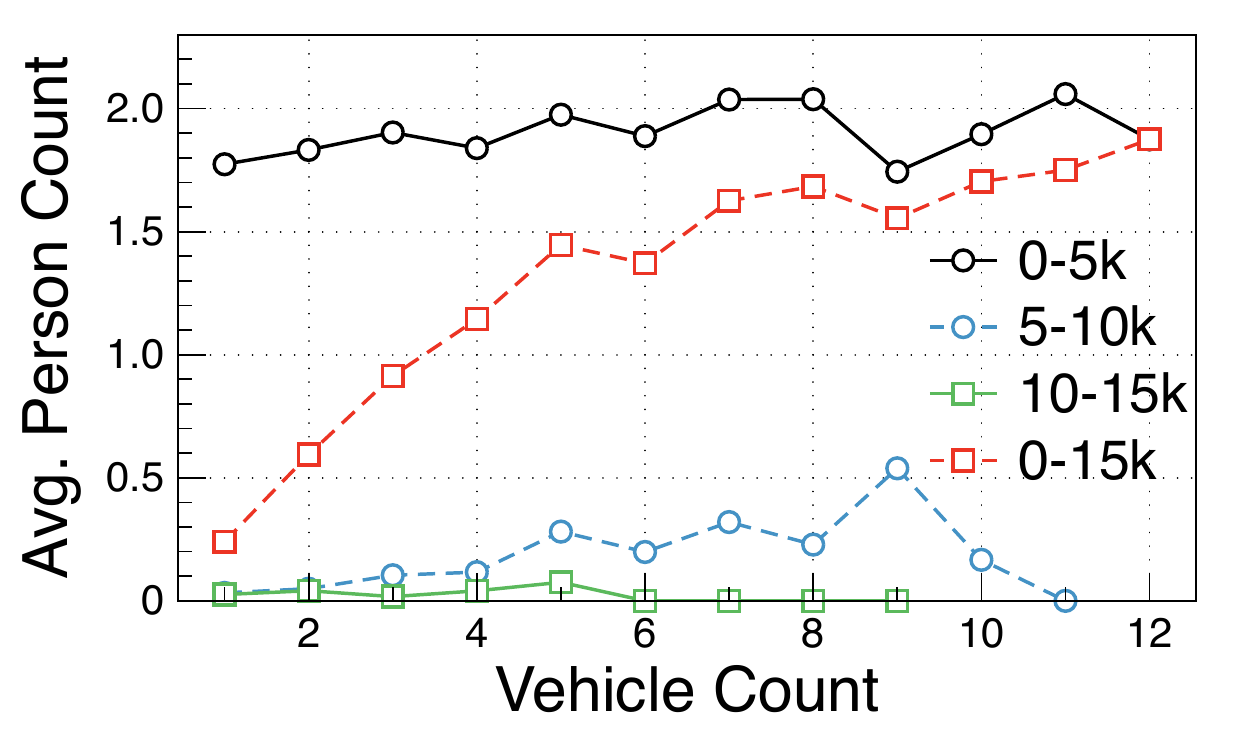}
         \caption{Vehicle Counting vs. Person Counting}
         \label{fig:motiv2p}
     \end{subfigure}
     \hfill
     \begin{subfigure}[b]{0.49\linewidth}
         \centering
         \includegraphics[width=\linewidth]{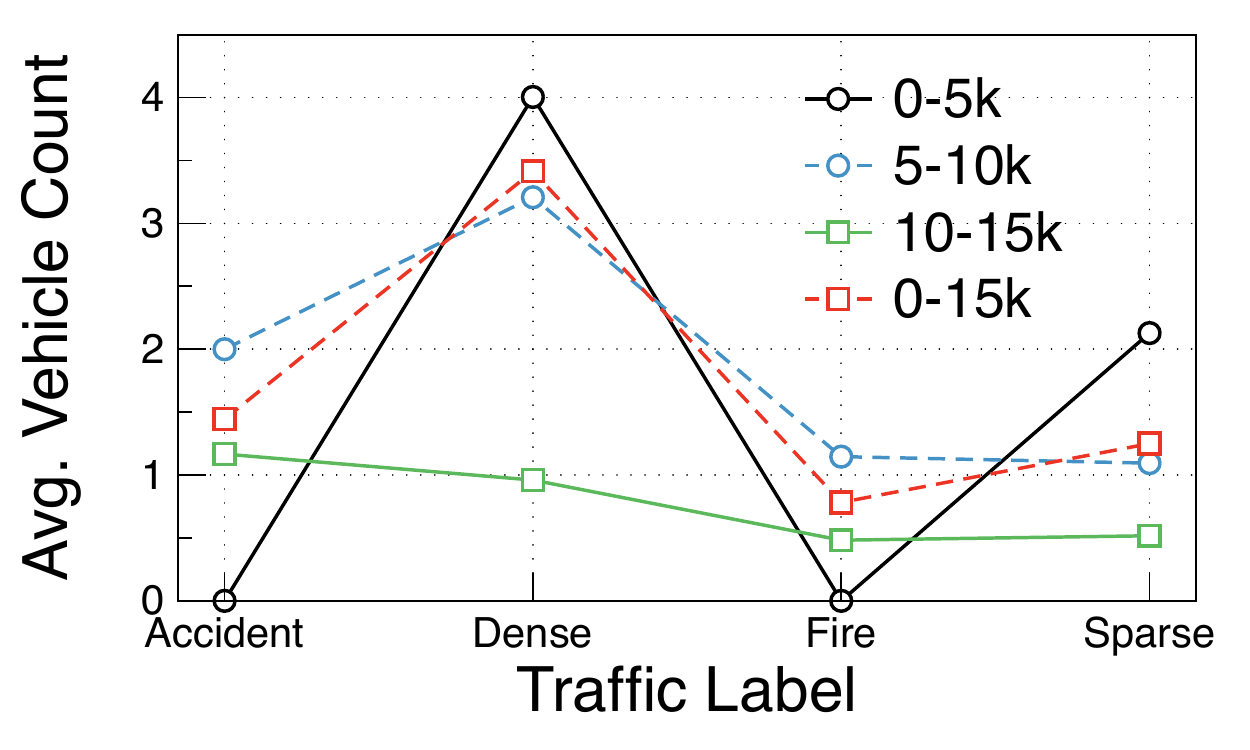}
         \caption{Traffic Condition vs. Vehicle Counting}
         \label{fig:motit2v}
     \end{subfigure}
\caption{Distribution shifts among different time periods. The legend $x$-$y$k means the video clip from the $x$k-th frame to the $y$k-th frame.}
\label{fig:online-moti}
\end{figure}

\subsection{Online Adaptive Training}

Typically, ML algorithms focus on the \textit{general} distribution of inputs.
In contrast, real-world inputs (e.g., the video stream captured by a camera) feature a much narrower distribution which changes on the fly~\cite{online-kd, li2021online}.
In a video analytics system (see Sec.~\ref{subsec:exp-setup} for detailed setup) where three models deployed (vehicle counting, person counting, and traffic condition classification), we analyzed the distribution shift over time.
As shown in Fig.~\ref{fig:motiv2p}, given the same vehicle count, there is a big difference between the four time periods.
And a similar difference also exists between the traffic condition classification and vehicle counting models (see Fig.~\ref{fig:motit2v}).
Thus, a vanilla approach that utilizes the initial outputs to train model links results in serious performance degradation when serving online.
And collecting samples that have a general enough distribution brings cold start problems.
Thus we study the online training approaches for model linking.

\textbf{Periodic update.}
A simple way to adapt to the dynamics of inputs is periodically collecting samples to update model links, e.g., run all ML models for 50 frames every ten minutes and train model links using the additional 50 samples.
Our experiments show that this approach is effective but lacks adaptability: it wastes computing resources to collect data during periods when updates are not necessary.
Therefore, we propose the following adaptive method that actively selects samples for model link updating.

\textbf{Adaptive update.}
Specifically, we use the uncertainty threshold method~\cite{active-learning-survey} to decide which data to label (i.e., running both source and target models to obtain the pair of matched inference results).
We utilize the canonical entropy uncertainty measurement that is applicable to both classification and regression models (the confidence is approximated by output variance)~\cite{regression-active}.
Besides uncertainty-based active policy, we adopt a loss prediction-based method~\cite{loss-pred}. 
The key idea is adding a neural network that takes the intermediate activation as input and predicts the sample loss. 
In principle, any active sampling approach is applicable to model linking.
Our experimental results show that, given the same labeling budget, online adaptive approaches can further improve the periodic one, due to its ability to sample at more necessary times.

\subsection{Linking for Domain Adaptation}

A long-standing problem of ML is that the model trained on the public dataset performs poorly on the target application data, namely domain shift.
Unsupervised~\cite{rangwani2022closer}, semi-supervised~\cite{singh2021clda}, and supervised~\cite{finetune} approaches have been comprehensively studied towards domain adaptation.
Under ground-truth supervision, our proposed model linking can also be used to adapt deployed black-box models to the target domain.

\textbf{Linking to an abstract model node.}
Under our model linking formalization, we first abstract a \textit{oracle} model node that represents the ground-truth generator (usually human annotators).
Next, we can build and train the model link from the deployed model (source domain) to the abstract model (target domain).
Then we can execute both the original model and the trained model link for serving.
Since we design the architecture of model links to be simple, a small collection of training samples are sufficient to obtain an effective model link.
The approach works with the insight that: unlike general ML that focuses on understanding a general distribution, serving models in real-world applications only needs to fit a much narrower distribution.

\begin{algorithm}[t]
\caption{Cross-Domain Aggregation}
\label{alg:aggregation}
\SetKwInOut{Input}{Input}
\Input{$K$ edges are indexed by $k$.}
\textbf{Cloud executes:}\\
Initialize global model link weights $\Theta_{G,0}$\;
\For{each round $t=1,2,...$}{
    \For{each edge $k$ \textbf{in parallel}}{
        $\Delta\Theta_{L,k}\leftarrow$ LocalUpdate($k$)\;
    }
    $\Theta_{G,t+1} \leftarrow \Theta_{G,t} + \frac{1}{K}\sum_{k} \Delta\Theta_{L,k}$\;
}

\textbf{Edge executes LocalUpdate($k$):}\\
Compute gradient vector $\Delta\Theta_{L,k}$ on local dataset\;
Upload $\Delta\Theta_{L,k}$ to the cloud\;
\end{algorithm}

\subsection{Cross-Domain Aggregation}

Let us consider a popular case where a cloud server holds ML models and deploys them to many edge devices for analyzing local data streams.
Then an interesting question for our proposed model linking is: how to aggregate model links that are locally trained in these edge devices?
Training model links avoids most privacy issues since it only needs the inference outputs of ML models, not requiring the raw sensory data.
So a \textit{global} model link can be trained by asking edges to send local inference results to the cloud server.
However, although inference results are usually less sensitive than the raw data, in some highly privacy-sensitive scenarios, e.g., medical images in hospitals~\cite{ng2021federated}, it is not allowed to transmit local analytic results.  
Following the idea of federated learning~\cite{fl-concept} which was proposed to train a global model with distributed and private data, we propose to train a global model link by aggregating gradients of corresponding local model links during training.
More specifically, let $g^G$ be a global model link with parameters $\Theta_G$ which is shared between local domains (e.g., edge devices).
Initially, the cloud server sends $g^G$ to every edge device.
In a round, each edge device trains local model link $g^L$ and sends gradients that are computed using local training samples to the cloud.
Then the server aggregates local changes by averaging the collected gradients and updates $\Theta_G$.
Algorithm.~\ref{alg:aggregation} shows the detailed procedures executed on the edges and cloud.
Apart from the benefit of better initialization, the global link can also be utilized to improve the domain adaptability~\cite{FL-DA} via fusing the local and global links by the same ensemble method presented in Sec.~\ref{subsec:ensemble}.
\section{Collaborative Multi-Model Inference}
\label{sec:inference}

In this section, we present a model link-based algorithm to schedule multi-model inference under a cost budget.

Let $\mathcal{F}(A)$ denote the average output accuracy, i.e.,
\begin{equation}
    \mathcal{F}(A)=\frac{1}{|F|}(\sum_{f_i \in A}1 + \sum_{f_j \in F\setminus A} p(A,f_j)).
\end{equation}
Then we define the gain of activating one more model $f_i$ as $\Delta(A, f_i)=\mathcal{F}(A\cup \{f_i\}) - \mathcal{F}(A)$.
Assuming that adding a source of model link into the ensemble model will not decrease the performance: $p(A\cup \{f_i\}, f_j) \geq p(A, f_j)$, which is empirically true~\cite{zhou2012ensemble}.
Then $\Delta(A, f_i) \geq 0$, i.e., the objective function is nondecreasing.
As for the submodularity, given $A_1 \subset A_2 \subset F, f_i\notin A_2$, we define $A_1'=A_1 \cup \{f_i\}, A_2'=A_2\cup \{f_i\}$.
Then we have: 
\begin{align*}
    \Delta(A_2, f_i) &- \Delta(A_1, f_i) = \frac{1}{|F|} \{ (p(A_1, j) - p(A_2, j))\\
    &+ \sum_{f_j\in A_2\setminus A_1, j\neq i}(p(A_1, f_j)-p(A_1', f_j)) \\
    +& \sum_{f_j\in F\setminus A_2, j\neq i}[\underbrace{(p(A_2',f_j)-p(A_2,f_j))}_{f_j \text{'s gain for } A_2} \\
    &- \underbrace{(p(A_1',f_j)-p(A_1,f_j))}_{f_j \text{'s gain for } A_1}]\}.
\end{align*}
Apparently, if the marginal gain of adding $f_j$ into $A_2, A_1$ is diminishing, then $\Delta(A_2,f_i)-\Delta(A_1,f_i)\leq 0$, i.e., the objective function is submodular.
But this property does not always hold.
In our experiments, we observed two typical cases:
(1) Dominance.
The performance of the ensemble model approximately equals the best-performance source of model links.
Let $f_{i^*} = argmax_{f_i \in A}p(g_{ij})$ denote the source with maximal performance.
We observe that $p(h_{A,f_j}) \approx p(g_{i^*j})$, i.e., the best source dominates the ensemble performance. 
(2) Mutual assistance.
The multi-source model links ensemble outperforms any single source.
$\forall f_i \in A, p(h_{A, f_j}) > p(g_{ij})$, i.e., sources of model links assist mutually.
And in this case, $f_j$'s gain for $A_2$ is possibly greater than its gain for $A_1, A_1 \subset A_2$, if $f_j$ collaborates better with models in $A_2 \setminus A_1$.

\begin{algorithm}[t]
\caption{Collaborative Multi-model Inference}
\label{alg:selection}
\SetKwInOut{Input}{Input}
\SetKwInOut{Output}{Output}
\Input{model set $F$,  cost budget $B$}
\Output{inference results $y_i$}
For every $f_i, f_j \in F, i\neq j$, train model links $g_{ij}$\;
For every $f_j \in F$, train ensemble model $h_{A_j, j}$, where $A_j=F\setminus \{f_j\}$\;
\For{each period}{
    Profile activation probability $\mathcal{P}_i$ for each $f_i \in F$ by Eq.~\ref{eq:activation}\;
    Greedily select $A \leftarrow A \cup \{argmax_{f_i \in F\setminus A}(\mathcal{P}_i)\}$ until reach the cost budget $B$\;
    Input $x$ arrives\;
    \For{$f_i \in F$}{
        \eIf{$f_i \in A$}{
            $y_i \leftarrow f_i(x)$\;
        }
        {
            $y_i \leftarrow h_{A,i}(\{y_j\}_{f_j \in A})$\;
        }
    }
}
\end{algorithm}

\begin{table*}[t!]
\centering
\caption{Summary of ML models used on Hollywood2 dataset.}
\begin{tabular}{@{}l|llll@{}}
\toprule
Task Class & ML Model & Input Modality & Output Format & Metric \\ \midrule
Single-Label Classification & Gender Classification~\cite{gender} & Audio & 2-D Softmax Labels & Accuracy \\ \midrule
Multi-Label Classification & Action Classification~\cite{c3d} & Video & 12-D Softmax Labels & mAP \\ \midrule
\multirow{2}{*}{Localization} & Face Detection~\cite{face} & Image & \multirow{2}{*}{4-D Bounding Boxes} & \multirow{2}{*}{IoU} \\
 & Person Detection~\cite{yolov3} & Image &  &  \\ \midrule
Regression & Age Prediction~\cite{age} & Image & 1-D Scalar & MAE \\ \midrule
\multirow{2}{*}{Sequence Generation} & Image Captioning~\cite{caption} & Image & \multirow{2}{*}{Variable-Length Text} & \multirow{2}{*}{WER} \\
 & Speech Recognition~\cite{speech} & Audio &  &  \\ \bottomrule
\end{tabular}
\label{tab:hollywood2-models}
\end{table*}

\begin{figure*}
     \centering
     \begin{subfigure}[b]{0.9\textwidth}
         \centering
         \includegraphics[width=\textwidth]{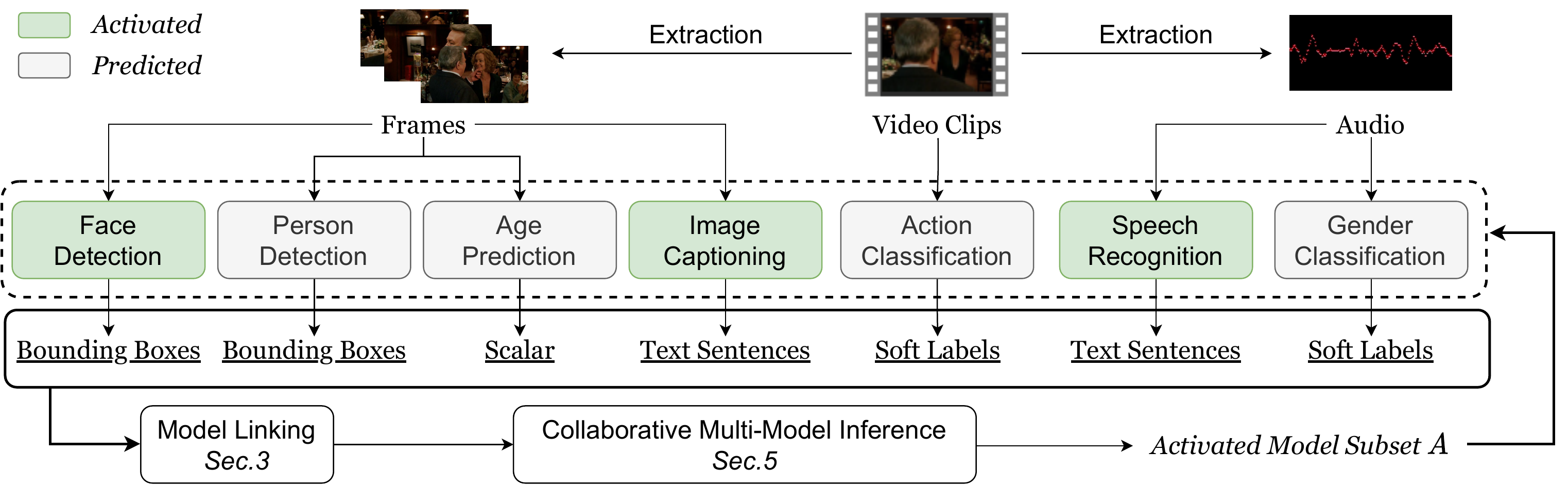}
     \end{subfigure}
\caption{Multi-task workflow on Hollywood2 dataset. It should be noted that the activated models are not fixed to these three models, but change dynamically.}
\label{fig:hw2-workflow}
\end{figure*}

\textbf{Activation probability.}
Solving Eq.~\eqref{eq:opt} is NP-hard and the ($1-e^{-1}$)-approximation algorithm~\cite{submodule-note} needs partial-enumeration and requires $O(n^5)$ computations of the objective function.
The optimization is not a one-off process and should be executed online to fit the dynamics of the inference system.
So we design a heuristic metric of activation probability, whose calculation only depends on the model links' performance rather than ensemble models'.
Given a model $f_i$, the activation probability considers three factors:
(1) the average performance of model links from $f_i$ to all the others, denoted by:
\begin{equation}
    \mathcal{P}_i^1=\frac{\sum_{j\neq i}p(g_{ij})}{|F|-1};
\end{equation}
(2) the average performance of model links targeted to $f_i$ from all the others, denoted by:
\begin{equation}
    \mathcal{P}_i^2=\frac{\sum_{j\neq i}p(g_{ji})}{|F|-1};
\end{equation}
(3) the cost of $f_i$, i.e., $c(f_i)$.
Then we design the activation probability as follows:
\begin{equation}
\label{eq:activation}
    \mathcal{P}_i=\frac{1+\mathcal{P}_i^1-\mathcal{P}_i^2}{wc(f_i)},
\end{equation}
where the weight $w$ can be determined by the following normalization. 
By regularizing the range into 0 to 1, we have $(1+1-0)/(w \min_i (c(f_i)) = 1$, i.e., $w = 2/ \min_i (c(f_i))$. 
This activation probability can be regarded as an coefficient that are positively correlated with the gain of the objective function when selecting a ML model.

\textbf{Periodic re-selection.}
Due to the content dynamics, the optimal subset of activated models may change over time.
But adapting to such dynamics brings additional overheads of loading and unloading ML models.
So we propose to periodically re-select activated models.
At the beginning of each period, we use a small proportion (e.g. 1\%) of the data for profiling the prediction performance of model links.
Then we update ML models' activation probabilities and re-select models to be loaded during the current period.
By reasonably setting the period length and the proportion of data used for profiling, we can amortize the overheads of loading/unloading ML models to negligible.

Algorithm.~\ref{alg:selection} shows the workflow of integrating MLinks with multi-model inference workloads.
Initially, we train pairwise model links and ensemble models.
During each period, we first calculate the activation probability by running all models on data for profiling.
Then we select greedily w.r.t. activation probability under the cost budget.
In the serving phase, activated models do exact inference while the others' outputs will be predicted by the model link ensemble of activated sources.
\section{Evaluation}
\label{sec:eval}

\subsection{Implementation}
We implemented our designs in Python based on TensorFlow 2.0~\cite{tf} as a pluggable middleware for inference systems~\footnote{https://github.com/yuanmu97/MLink}.
We tested the integration on programs implemented with TensorFlow~\cite{tf}, PyTorch~\cite{pytorch} and MindSpore~\cite{ms}, with only dozens of lines of code modification, which shows the ease of use of \textit{MLink}.

\subsection{Experiment Setup}
\label{subsec:exp-setup}
We evaluated our designs on a multi-modal dataset and two real-world video analytics systems.

\textbf{Multi-modal dataset and ML models.}
We used the Hollywood2 video dataset~\cite{hollywood2}.
To obtain aligned inputs for multi-modal models, we selected the 30th frame and extracted audio data from each video. 
We deployed seven pre-trained ML models that cover five classes of learning tasks: single-label and multi-label classification, object localization, regression, and sequence generation.
And they have different model architectures, input modalities and output formats.
To evaluate the performance of model links, we used task-specific metrics, including accuracy, mean average precision (mAP), intersection over union (IoU) of the bounding box, mean absolute error (MAE), and word error rate (WER).
Tab.~\ref{tab:hollywood2-models} summarizes information of these ML models.

\textbf{Smart building and city traffic monitoring systems.}
We evaluated \textit{MLink} on two real-world video analytics systems.
1) Smart building.
To support applications including automatic air conditioning and lighting, abnormal event monitoring, and property security, three ML models were deployed: OpenPose~\cite{openpose}-based person counting, ResNet50~\cite{resnet}-based action classification~\cite{actionet}, and YOLOV3~\cite{yolov3}-based object counting.
We collected two days (one weekday and one weekend) of video frames from all 58 cameras (1 frame per minute).
We use an edge server with one NVIDIA 2080Ti GPU.
2) Traffic monitoring.
On a city-scale video analytics platform with over 20,000 cameras, three AI models were deployed for traffic monitoring: OpenPose~\cite{openpose}-based person counting, ResNet50~\cite{resnet}-based traffic condition classification~\cite{traffic}, and YOLOV3~\cite{yolov3}-based vehicle counting.
We selected 10 cameras at the road intersections and collected two days (one weekday and one weekend) of frames (1 FPS).
We used five servers, each with four NVIDIA T4 GPUs.

\textbf{Baselines.}
We introduce the naive standalone inference and three strong alternative resource-performance trade-off approaches as baselines.
(1) \textbf{Standalone}: running models independently.
(2) \textbf{MTL}: We adopt a multi-task learning architecture~\cite{mtl-survey} that consists of a global feature extractor shared by all tasks and task-specific output branches.
We use ResNet50~\cite{resnet} to implement the feature extractor and fully-connected layers for task-specific outputs.
We initialize the ResNet50 feature extractor with weights pretrained on ImageNet~\cite{imagenet} and connect three output branches for person counting, action/traffic classification, and object/vehicle counting tasks, on smart building/traffic monitoring testbeds.
The MTL models are trained under the supervision of exact inference results of corresponding models.
(3) \textbf{Reducto}~\cite{reducto}: a frame filtering approach.
For each model, Reducto first computes the feature difference of successive frames.
If the feature difference is lower than a threshold, it filters out the current frame and reuses the latest inference output.
We tested four types of low-level features as proposed in Reducto and selected the one that has the best performance.
(4) \textbf{DRLS} (Deep Reinforcement Learning-based Scheduler)~\cite{adams}: a multi-model scheduling approach.
DRLS trains a deep reinforcement learning agent to predict the next model to execute on the given data, based on the observation of executed models' outputs.

\subsection{Black-Box Model Linking}
\label{subsec:exp-link}

\begin{figure}
     \centering
     \begin{subfigure}[b]{0.22\textwidth}
         \centering
         \includegraphics[width=\textwidth]{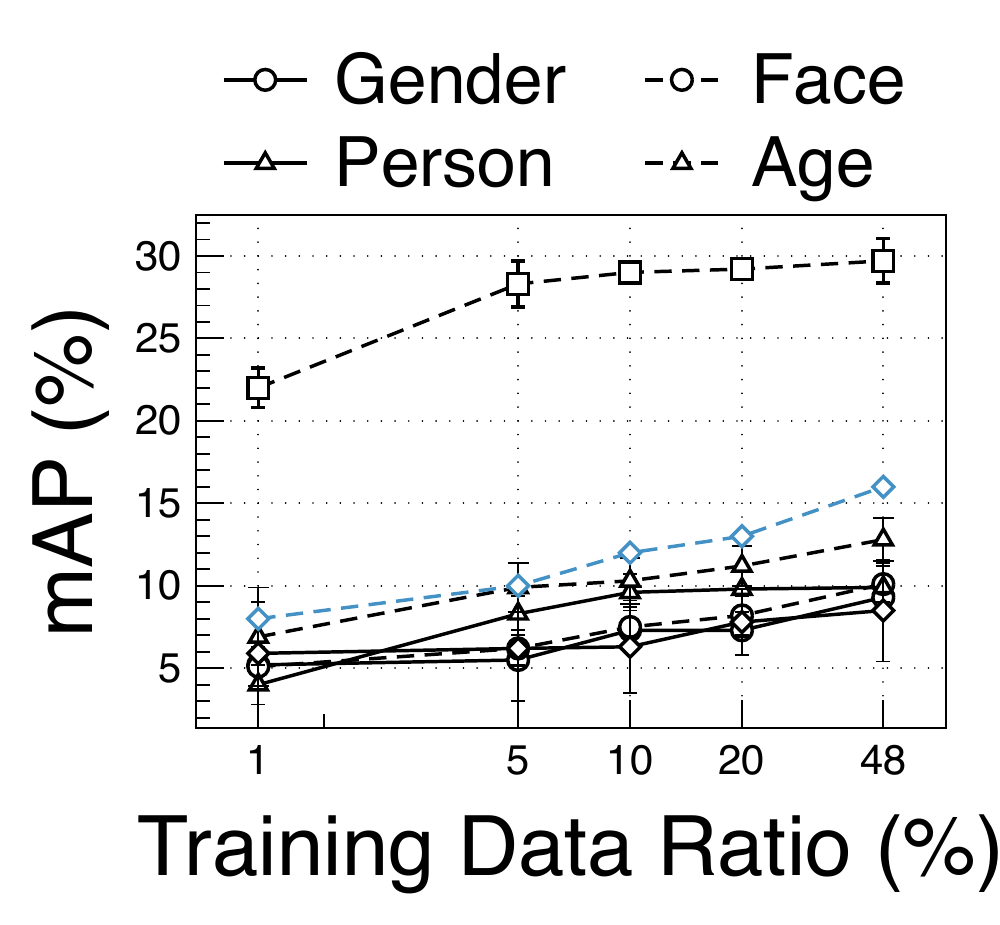}
         \caption{Target: Action}
         \label{fig:gender}
     \end{subfigure}
     \begin{subfigure}[b]{0.22\textwidth}
         \centering
         \includegraphics[width=\textwidth]{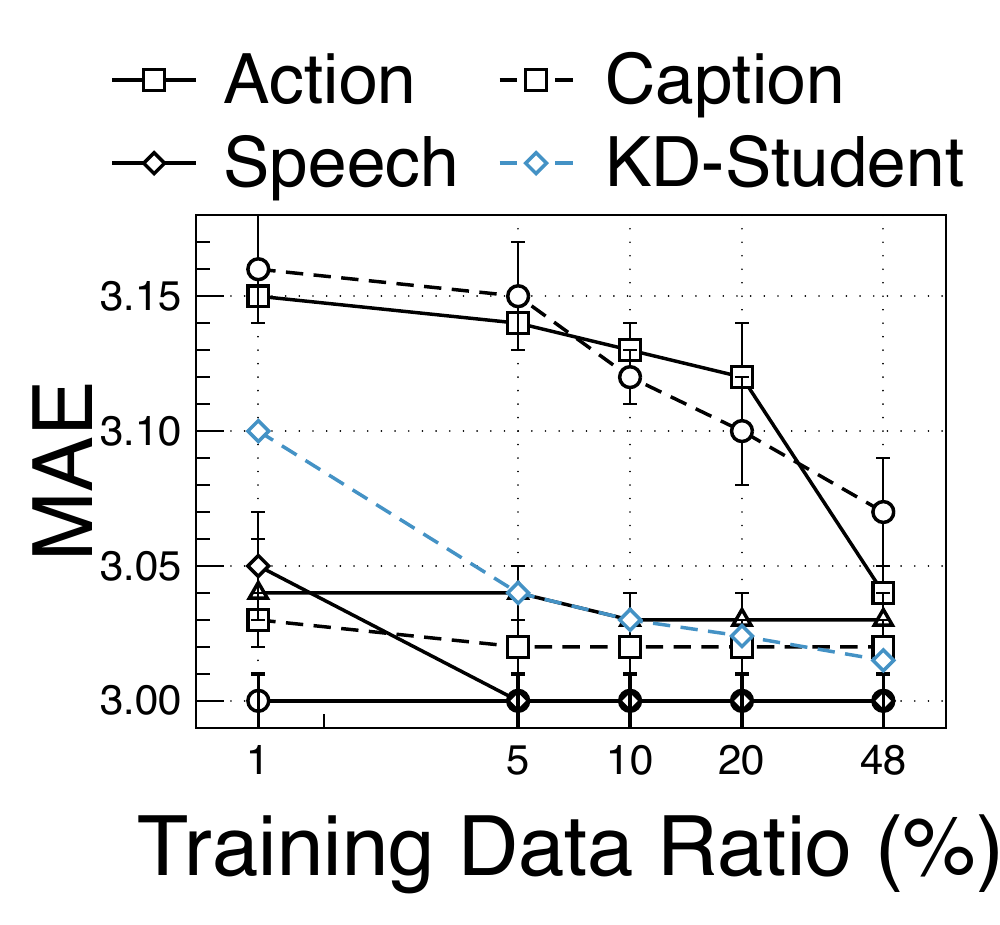}
         \caption{Target: Age}
         \label{fig:face}
     \end{subfigure}
     \begin{subfigure}[b]{0.22\textwidth}
         \centering
         \includegraphics[width=\textwidth]{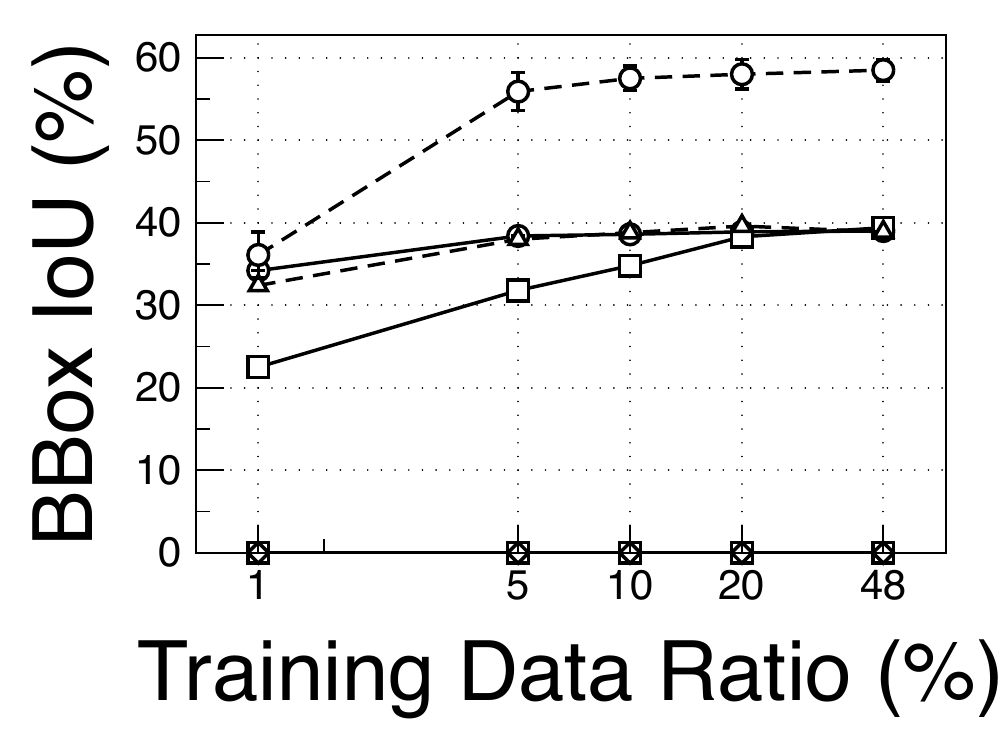}
         \caption{Target: Person}
         \label{fig:person}
     \end{subfigure}
     \begin{subfigure}[b]{0.22\textwidth}
         \centering
         \includegraphics[width=\textwidth]{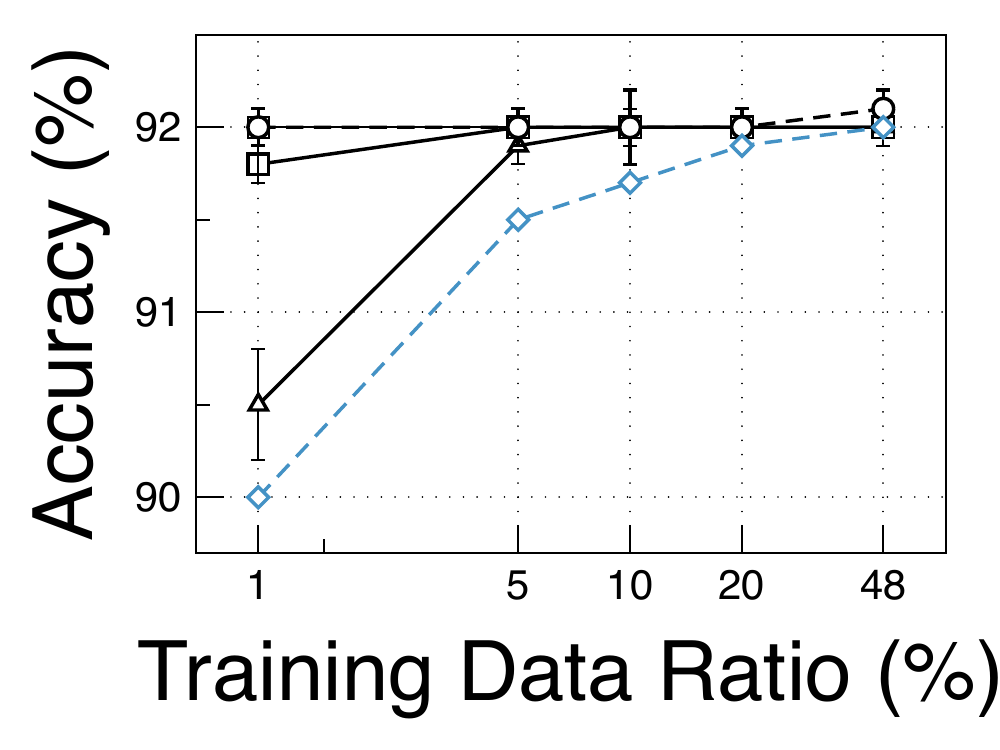}
         \caption{Target: Gender}
         \label{fig:age}
     \end{subfigure}
\caption{Performance of model links from different source models on four targets. KD-Student: student model trained via knowledge distillation on the target model.}
\label{fig:hw2-mlink}
\end{figure}

\textbf{Sensitivity to the size of training data.}
The original training and test splits in Hollywood2 dataset contain 823 video clips (around 48\%) and 884 video clips, respectively.
To test the performance of the model linking with different sizes of training data, we further randomly sampled four subsets of training data with 1\%, 5\%, 10\%, 20\% ratios, with respect to the total dataset.
We trained pairwise model links with the RMSprop~\cite{rmsprop} optimizer and the same hyper-parameters (0.01 learning rate, 100 epochs, 32 batch size).
As a fair comparison, we adopt a knowledge distillation~\cite{kd} method for some target models (Action, Age, Gender) where the student model has two convolutional layers. 
We repeated the experiments three times and reported the mean and standard deviation of performance.
As shown in Fig.~\ref{fig:hw2-mlink}, using all training data, the \textit{Caption}-to-\textit{Action} model link can achieve 31.7\% mAP.
And the model links between the two detection models, \textit{Face}-to-\textit{Person} and \textit{Person}-to-\textit{Face} model links, achieve 59\% and 32\% IoU, respectively.
Even with very limited training samples, 1\%, some model links achieve high performance.
The model link from \textit{Face} to \textit{Gender} achieves 92.1\% accuracy.
And for model links \textit{Gender}-to-\textit{Age} model achieves 3.0 MAE.
Compared with student models trained via knowledge distillation on the target models, model links achieves higher prediction performance, especially when the amount of training data is small.
But for speech recognition and video caption models, model links targeted to them cannot be effectively built and have around one WER score.

\begin{table}[t]
\centering
\caption{IoU scores of model links targeted to the \textit{Person} model and the Pearson correlations.}
\begin{tabular}{l|cccc}
\toprule
Source & Action & Age & Face & Gender \\ \midrule
IoU (\%) & 39.4 \tiny{($\pm$0.1)} & 38.9 \tiny{($\pm$0.1)} & \textbf{58.5} \tiny{($\pm$1.3)} & 39.0 \tiny{($\pm$0.1)} \\
Corr. & 0.123 & 0.042 & \textbf{0.244} & -0.053 \\ \bottomrule
\end{tabular}
\label{tab:corr}
\end{table}

\textbf{Model link ensemble.}
For one target model, we have built multiple model links from different source models.
Then we trained the ensemble models with all sources, using the same optimizer as model links and same hyper-parameters.
Tab.~\ref{tab:ensemble} shows the results on five target models, both model links and ensemble models were trained using all training samples (48\% ratio).
The model link ensemble outperforms every single source model.
We can see there are two typical cases: dominance and mutual assistance.
For \textit{Action}, \textit{Face}, \textit{Person} targets, the \textit{Caption}, \textit{Person}, \textit{Face} sources dominate the ensemble performance, respectively.
But for \textit{Age} and \textit{Gender} targets, source models mutually assist and achieve performance improvement by the ensemble.

\textbf{Correlation quantification.}
We calculated the Pearson correlation coefficients between inference outputs of different models on the training split.
For single-label and multi-label classification models, we used the index with the highest confidence as the label.
For localization models, we checked whether the bounding box is empty, and assign 0 or 1 as the label.
We used the regression scalar as the label and skipped the two sequence generation models.
Tab.~\ref{tab:corr} shows the results of model links targeted to the \textit{Person} model, and we can see a positive correlation between the model link performance and the Pearson correlation coefficient.

\textbf{Discussions.}
To explore the limitations of MLink, we consider cross-domain semantic segmentation tasks on Cityscapes~\cite{cityscapes} and GTAV~\cite{gtav} datasets. 
We deployed the DeepLabV3Plus~\cite{deeplabv3plus} model pre-trained on Cityscapes and ran it on GTAV images. 
Then we trained model links that map model predictions to ground-truth masks. 
Experimental results show that model linking is not effective in this case: the accuracy of remapped predictions degrades by around 10\%. 
The first reason is the large output space of pixel-level segmentation tasks. Learning to calibrate predictions on two million (1052*1914) pixels is difficult. 
The second reason is that our vec-to-vec design (fully-connected layers) is task-agnostic. We did not use down-sampling and up-sampling convolutional neural network architecture, which is the best practice for the semantic segmentation task.

\begin{table*}[t]
\centering
\caption{Dominance and mutual assistance cases in model link ensemble. Column titles are source models and row titles are target models. The dominant source's performance is in bold.}
\begin{tabular}{@{}l|ccccccc|c@{}}
\toprule
Target $\backslash$ Source & Action & Age & Caption & Face & Gender & Person & Speech & Ensemble \\ \midrule
Action mAP(\%) & - & 12.8($\pm$1.3) & \textbf{29.7($\pm$1.4)} & 10.1($\pm$1.3) & 9.3($\pm$0.3) & 9.9($\pm$1.2) & 8.5($\pm$3.1) & 30.8($\pm$1.1) \\
Face IoU(\%) & 11($\pm$1.3) & 11.2($\pm$1.0) & 0 ($\pm$0) & - & 10.3($\pm$0.9) & \textbf{31.9($\pm$0.3)} & 0 ($\pm$0) & 32.2($\pm$0.2) \\
Person IoU(\%) & 39.4($\pm$0.1) & 38.9($\pm$0.1) & 0($\pm$0) & \textbf{58.5($\pm$1.3)} & 39.0($\pm$0.1) & - & 0($\pm$0) & 59.2($\pm$1.2) \\ \midrule
Age MAE & 3.04($\pm$0.01) & - & 3.02($\pm$0.01) & 3.07($\pm$0.02) & 3.0($\pm$0.01) & 3.03($\pm$0.01) &  3.0($\pm$0.01) & 2.98($\pm$0) \\
Gender Acc.(\%) & 92($\pm$0.1) & 92.1($\pm$0.2) & 92($\pm$0.1) & 92.1($\pm$0.1) & - & 92($\pm$0.1) & 92($\pm$0.1) & 92.3($\pm$0)\\ \bottomrule
\end{tabular}
\label{tab:ensemble}
\end{table*}

\subsection{Online MLink Training}

\begin{figure}[t]
    \centering
    \includegraphics[width=0.9\linewidth]{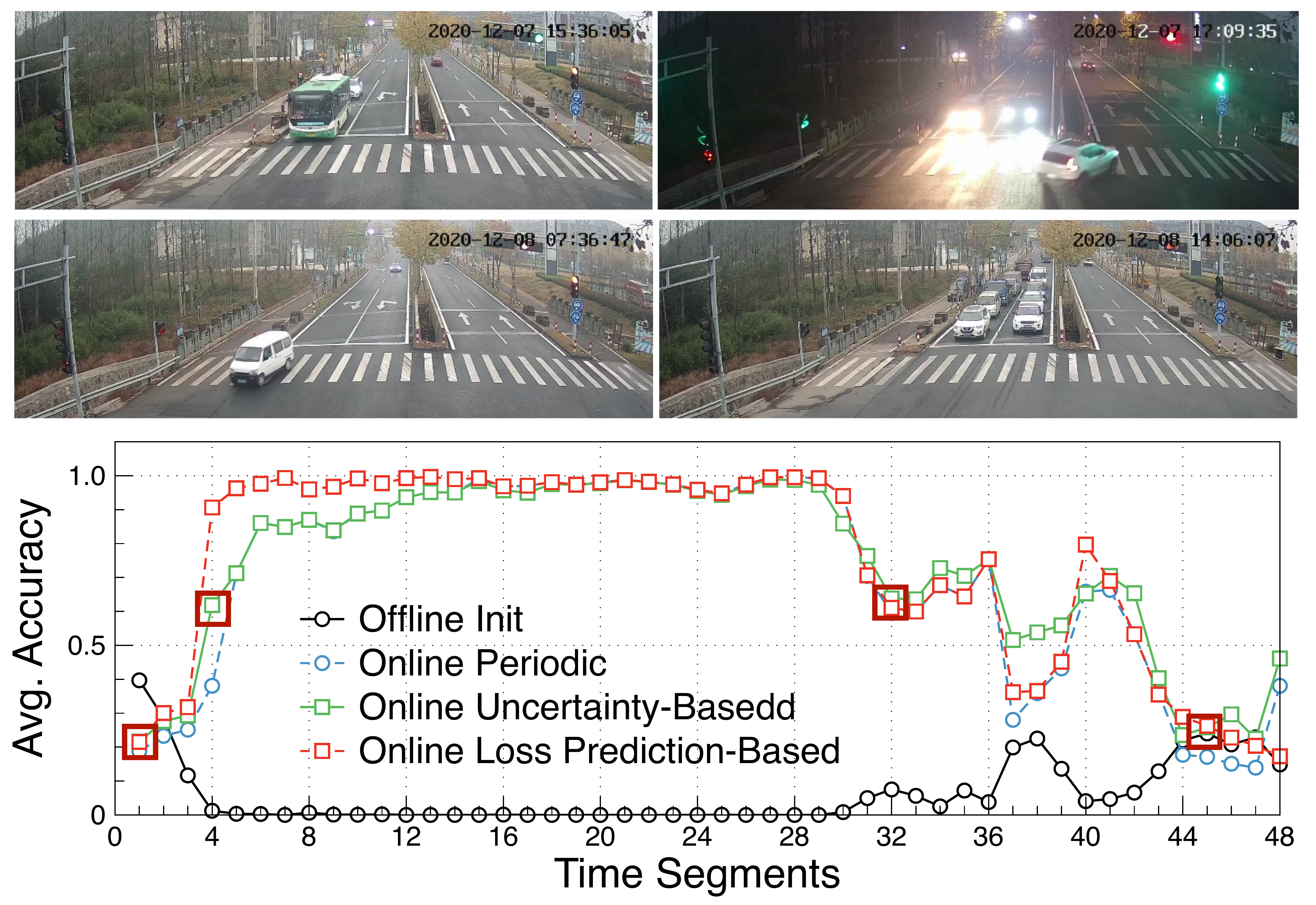}
\caption{Online Training. Vehicle-to-Person MLink. Images correspond to boxed points in red and show times where the distribution significantly changed.}
\label{fig:online}
\end{figure}

We evaluated our proposed online training approach for model linking on the traffic monitoring application.
For the model link from the vehicle counting source model to the person counting target model, Fig.~\ref{fig:online} shows the segment-level (one segment per hour) accuracy on one camera of different training approaches.
We set the ratio of training samples as 1\%.
The \textit{Offline Init} approach uses the first 1\% samples for training the model link and does not update it for the following data.
Experimental results show that due to the limited distribution of training samples, \textit{Offline Init} approach returns $<$1\% accuracy on 26 segments and 6.3\% accuracy on average.
Our proposed \textit{Online Periodic} significantly improves the average accuracy to 70.2\% and \textit{Online Uncertainty-Based} brings an additional 3.3\% improvement.
Our loss prediction-based approach outperforms the others and achieves 74.7\% average accuracy.

\subsection{MLink Adaptation and Aggregation}
\textbf{Domain adaptation on a public dataset.}
We used Office-Home dataset~\cite{office-home}  to evaluate the domain adaptation performance of model links.
The dataset contains images on four domains: Art (A), Clip Art (C), Product (P) and Real World (R).
The splits of training and test data contain 7728 and 7860 images, respectively.
As the baseline, we trained ResNet50~\cite{resnet} classifiers separately by all training samples in each domain.
Tested on the same domain (marked as \textit{Source}), these models achieve 87.9\%, 89.9\%, 96.7\%, and 95.0\% accuracy scores, respectively.
When adopting these model to different domains (marked as \textit{Shift}), they suffer significant performance degradation due to domain shift and 
the average accuracy drops severely from 92.3\% to 59.5\%.
Using 10\% randomly sampled data from the training split in the target domain, we trained model links from ResNet50 in the source domain to the target domain.
For comparison, we tested two state-of-the-art approaches of domain adaptation: an unsupervised SDAT~\cite{rangwani2022closer} and a semi-supervised CLDA~\cite{singh2021clda}.
The detailed results are shown in Tab.~\ref{tab:office-home}.
Using all target-domain images in the unsupervised / semi-supervised way, SDAT / CLDA improves the average accuracy to 72.2\% / 75.5\%.
Using only 10\% of target-domain training samples and treating ResNet50 as a black box, model links achieve 72.6\% average accuracy. 
The results show that model links can cost-effectively mitigate the effect of domain shift and improve the adaptability of black-box ML models.

\begin{table*}[t]
\centering
\caption{Domain adaptation performance comparison on Office-Home. MLink achieves higher mean accuracy than unsupervised approach (SDAT) and comparable performance with the semi-supervised one (CLDA).}
\label{tab:office-home}
\begin{tabular}{@{}l|ccc|ccc|ccc|ccc|c@{}}
\toprule
Method & \multicolumn{1}{l}{A-\textgreater{}C} & \multicolumn{1}{l}{A-\textgreater{}P} & \multicolumn{1}{l|}{A-\textgreater{}R} & \multicolumn{1}{l}{C-\textgreater{}A} & \multicolumn{1}{l}{C-\textgreater{}P} & \multicolumn{1}{l|}{C-\textgreater{}R} & \multicolumn{1}{l}{P-\textgreater{}A} & \multicolumn{1}{l}{P-\textgreater{}C} & \multicolumn{1}{l|}{P-\textgreater{}R} & \multicolumn{1}{l}{R-\textgreater{}A} & \multicolumn{1}{l}{R-\textgreater{}C} & \multicolumn{1}{l|}{R-\textgreater{}P} & \multicolumn{1}{l}{Avg.} \\ \midrule
Source & \multicolumn{3}{c|}{87.9} & \multicolumn{3}{c|}{89.9} & \multicolumn{3}{c|}{96.7} & \multicolumn{3}{c|}{95.0} & 92.3 \\
Shift & 46.9 & 65.2 & 71.8 & 49.0 & 60.9 & 63.5 & 53.6 & 44.1 & 73.3 & 61.9 & 47.9 & 76.1 & 59.5 \\
SDAT~\cite{rangwani2022closer} & 58.2 & 77.1 & 82.2 & 66.3 & 77.6 & 76.8 & 63.3 & 57.0 & 82.2 & 74.9 & 64.7 & 86.0 & 72.2 \\
CLDA~\cite{singh2021clda} & 63.4 & 81.4 & 81.3 & 70.5 & 80.9 & 80.3 & 72.4 & 63.9 & 82.2 & 76.7 & 66.0 & 87.6 & 75.5 \\ \midrule
MLink & 69.2 & 85.3 & 80.3 & 59.8 & 77.1 & 71.4 & 62.9 & 64.2 & 80.1 & 69.7 & 66.1 & 86.1 & 72.6 \\ \bottomrule
\end{tabular}
\end{table*}

\begin{figure}
     \centering
     \begin{subfigure}[b]{0.24\textwidth}
         \centering
         \includegraphics[width=\textwidth]{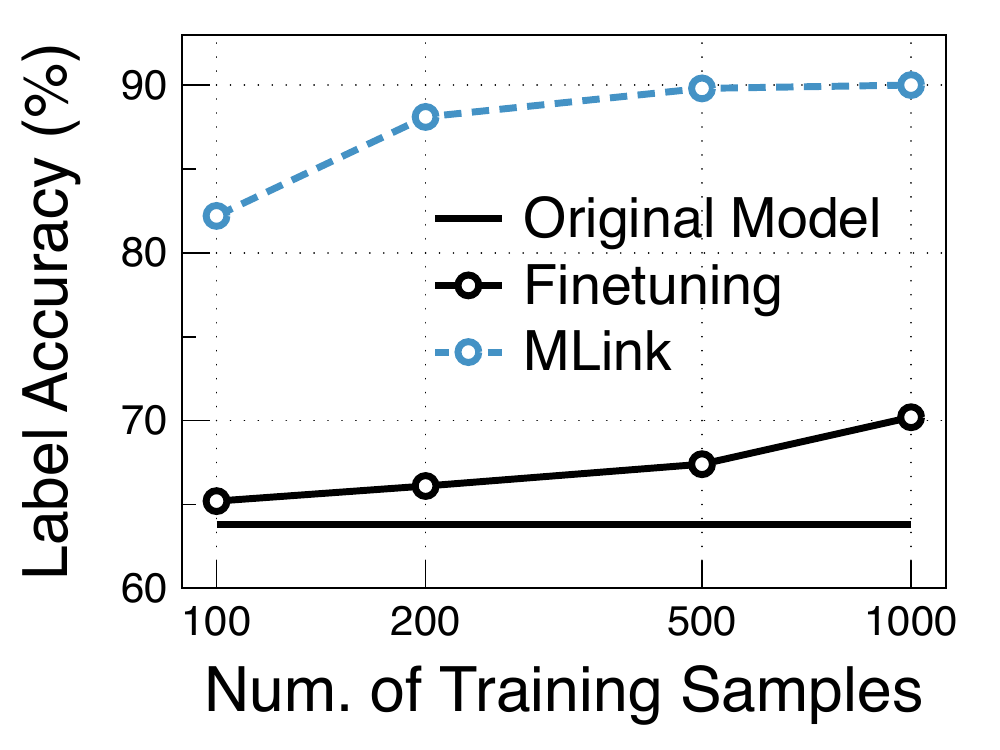}
         \caption{Gym Domain}
         \label{fig:gymweekend}
     \end{subfigure}
     \hfill
     \begin{subfigure}[b]{0.24\textwidth}
         \centering
         \includegraphics[width=\textwidth]{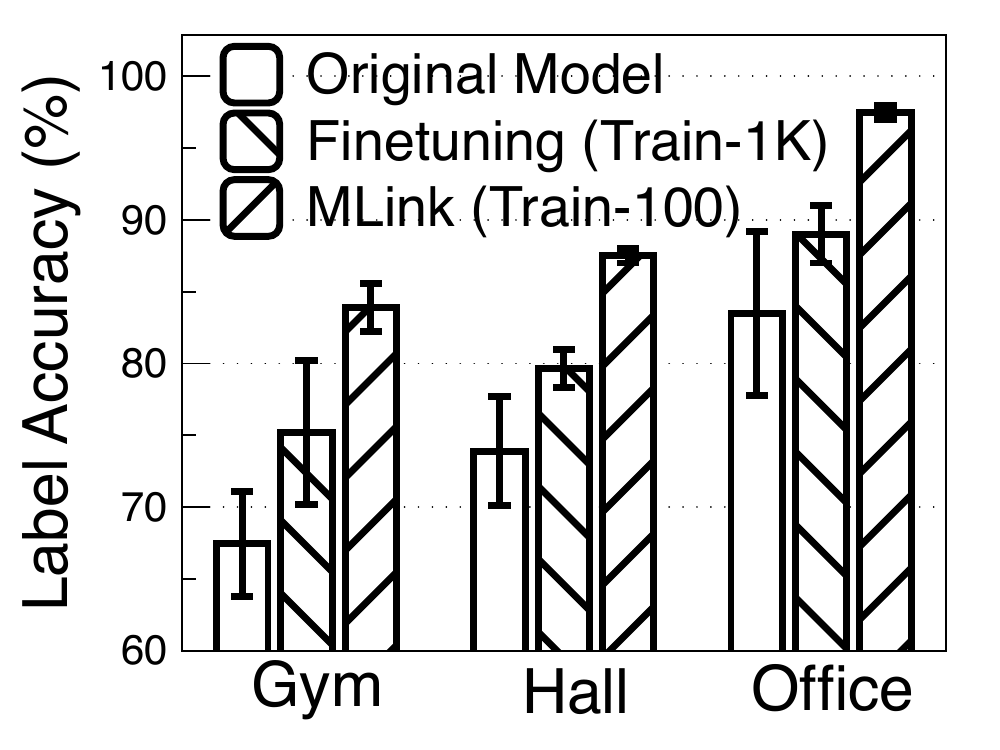}
         \caption{Smart Building Overall}
         \label{fig:da-build-all}
     \end{subfigure}
\caption{Domain adaptation performance on smart building. Three domains.}
\label{fig:da-smartbuild}
\end{figure}

\textbf{Domain adaptation on real-world applications.}
As introduced in Sec.\ref{subsec:exp-setup}, the three models developed for the smart building application were initially trained by public datasets.
Among them, the action classifier suffered the most serious performance degradation when applied to the videos captured in the building.
We divide the 58 cameras into three real scenarios: \textit{Gym} (G), \textit{Hall} (H), and \textit{Office} (O), with 11, 25, and 22 cameras, respectively.
And we collected 2000 images in each scenario and manually labeled the human actions.
The labeled data were split into training and test subsets with 1000 images for each.
The test accuracy of the pre-trained action classifier in \textit{Gym}, \textit{Hall}, and \textit{Office} are only 67.5\%, 73.9\%, and 83.5\%, respectively.
Using the pre-trained action classifier's outputs and the ground-truth labels, we trained model links using the different number of samples (100, 200, 500, 1000).
In comparison, we adopt a classifier fine-tuning method~\cite{finetune-bestpractice} that freezes parameters of the feature extractor in the pre-trained model and retrains the classifier.
Fig.~\ref{fig:gymweekend} plots the accuracy tested on the \textit{Gym} domain, where fine-tuned classifier achieves 66.1\% accuracy while the model link significantly outperforms it with 88.1\% accuracy using only 200 training samples. 
As presented in Fig.~\ref{fig:da-build-all}, with 90\% fewer training samples, model links still increase the label accuracy by up to 20.7\% and outperform fine-tuned classifier by at least 7.85\% on average accuracy improvement of the pre-trained model.
The reason is that fine-tuning works in high-dimensional feature space, which is so complex that limited training samples cannot tune it to fit the target domain.
But model links focus on the adaptation relations that have much lower dimensions.

\begin{figure}
     \centering
     \begin{subfigure}[b]{0.24\textwidth}
         \centering
         \includegraphics[width=\textwidth]{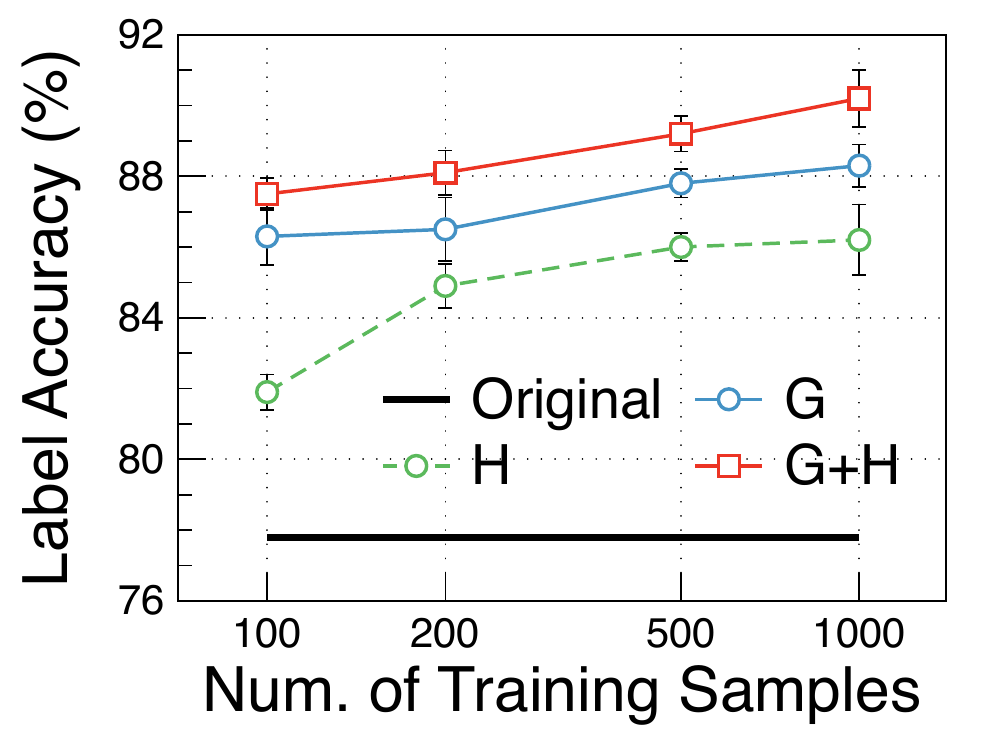}
         \caption{Model Link Aggregation Tested on \textit{Office} Domain}
         \label{fig:aggregation}
     \end{subfigure}
     \hfill
     \begin{subfigure}[b]{0.24\textwidth}
         \centering
         \includegraphics[width=\textwidth]{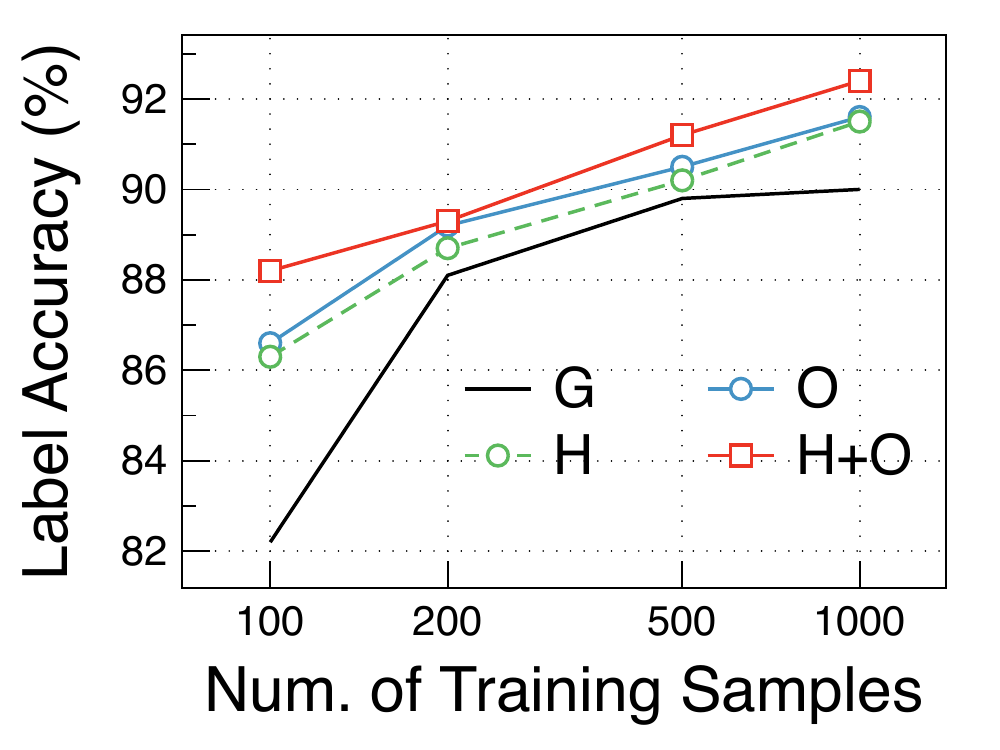}
         \caption{Local-Global Fusion Tested on \textit{Gym} Domain}
         \label{fig:lg-fusion}
     \end{subfigure}
\caption{Model link aggregation and fusion performance on smart building with three domains (G for \textit{Gym}, H for \textit{Hall}, and O for \textit{Office}).}
\label{fig:if-smartbuild}
\end{figure}

\textbf{Cross-domain aggregation.}
We used three smart building scenarios to evaluate the global model links trained by aggregating cross-domain local model links.
Taking one scenario as the target domain and the other two as the source domains, we aggregate local model links trained on source domains into the global model link.
For the action classification, Fig.~\ref{fig:aggregation} plots the label accuracy of local and global model links in the \textit{Office} scenario using different number of samples for training.
Directly applying the local model link trained in \textit{Gym} (or \textit{Hall}) to \textit{Office} can improve the accuracy of the original model from 77.8\% to 86.2\% (or 88.3\%).
By applying the global model link that was aggregated from \textit{Gym} and \textit{Hall} (G+H), the label accuracy achieves 90.2\%.
On average, the global model links from the other two domains bring a 7.85\% improvement in accuracy to the target domain, without using any sample in the target domain.

\textbf{Local-global fusion.}
We evaluated the effect of fusing local and global model links on the smart building system.
For the action classification model, Fig.~\ref{fig:lg-fusion} shows the accuracy tested on the \textit{Gym} domain from which we can see that fusing the local model link (marked as G) and other domains (O, H, O+H) improves the accuracy up to 6\% accuracy than only using the \textit{Gym} domain data.
For overall results on all three domains, on average, aggregating cross-domain model links brings an additional 1.1\% improvement in accuracy.

\begin{figure}[t]
    \centering
    \includegraphics[width=0.9\linewidth]{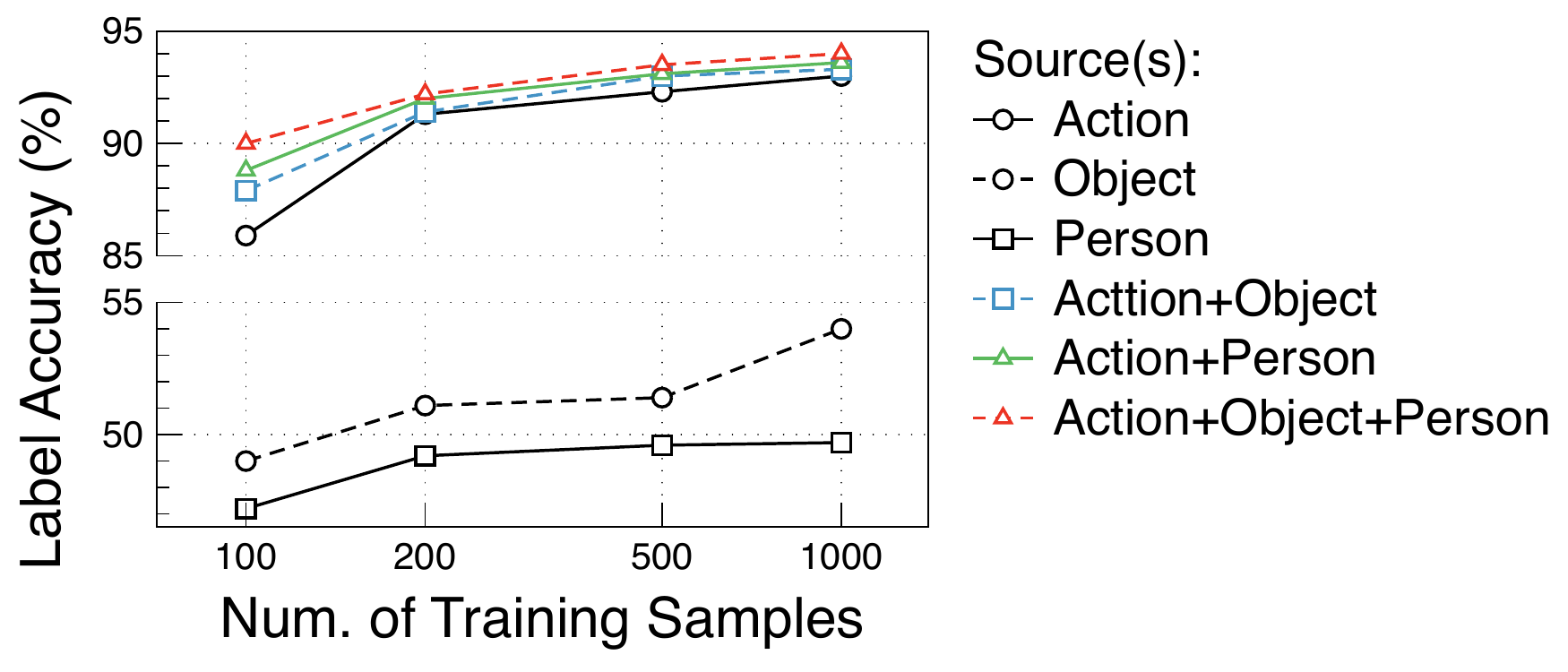}
\caption{Cross-task model link fusion on the smart building application with Action as the target task.}
\label{fig:crosstask}
\end{figure}

\textbf{Cross-task fusion.}
Then we trained model links from the other two models, object counting (Object) and person counting (Person) models, to the action classifier (Action).
And we tested the impact of fusing these links sourced from cross-task models.
Fig.~\ref{fig:crosstask} plots the label accuracy of model links with fused weights trained by the different number of training samples in the smart building application.
The model links fused from multiple sources significantly outperform the single-source model links. 
Compared with the ``Action'' model link, fusing links from three cross-task models can improve the accuracy by up to 4.1\%.

\subsection{Video Analytics with Model Links}

We test \textit{MLink} on 48-hour videos of 58 cameras in a smart building system and 48-hour videos of 10 cameras on a city traffic monitoring platform.
We leveraged the first 10\% in time of data for training model links and ensembles.
We set the period length as one hour and use initial 1\% data for profiling activation probabilities.
For the counting models, the output accuracy is calculated by checking whether the absolute error of the predicted number is less than 0.5.
The time costs of each ML model were the average inference time offline profiled by the training data.
In the smart building system, the action/person/object models cost 30/44/60 ms per frame.
In the traffic monitoring system, the traffic/person/vehicle models cost 55/66/70 ms per frame.
The GPU memory costs of each ML model were the peak usage: 4.6 GB for person counting, 1.5 GB for action/traffic classification, and 3.7 GB for object/vehicle counting.
We set the budget $B$ as the maximal GPU memory allocated for ML models to evaluate how \textit{MLink} improves the resource efficiency of multi-model inference.
We treat every ML model's output accuracy equally and report their average output accuracy.
Under GPU memory budget, the baseline ``Standalone'' simply selects the model with minimal average time cost.
We repeated the scheduling experiments three times and reported the results in Tab.~\ref{tab:schedule}.
Since the standard deviations are small ($< 0.1$), we did not present them for simplicity.
In both systems, \textit{MLink} outperforms alternatives in output accuracy.
Compared with ``Standalone'', in the smart building system, \textit{MLink} saves 66.7\% inference executions, while preserving 94.1\% output accuracy.

\begin{figure}
     \centering
     \begin{subfigure}[b]{0.24\textwidth}
         \centering
         \includegraphics[width=\textwidth]{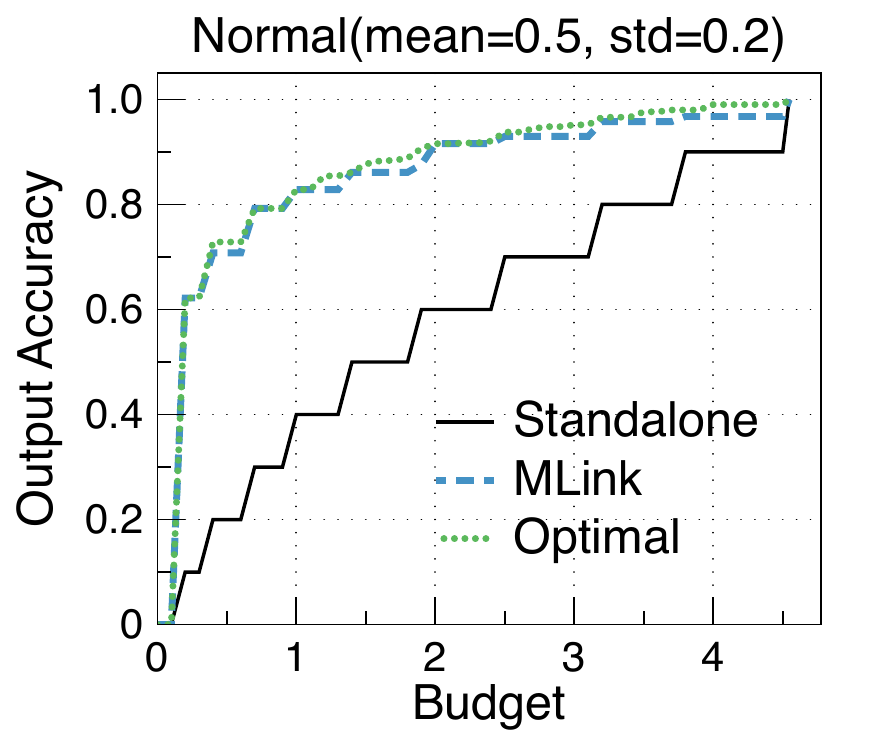}
         \caption{Normal Distribution}
         \label{fig:normal}
     \end{subfigure}
     \hfill
     \begin{subfigure}[b]{0.24\textwidth}
         \centering
         \includegraphics[width=\textwidth]{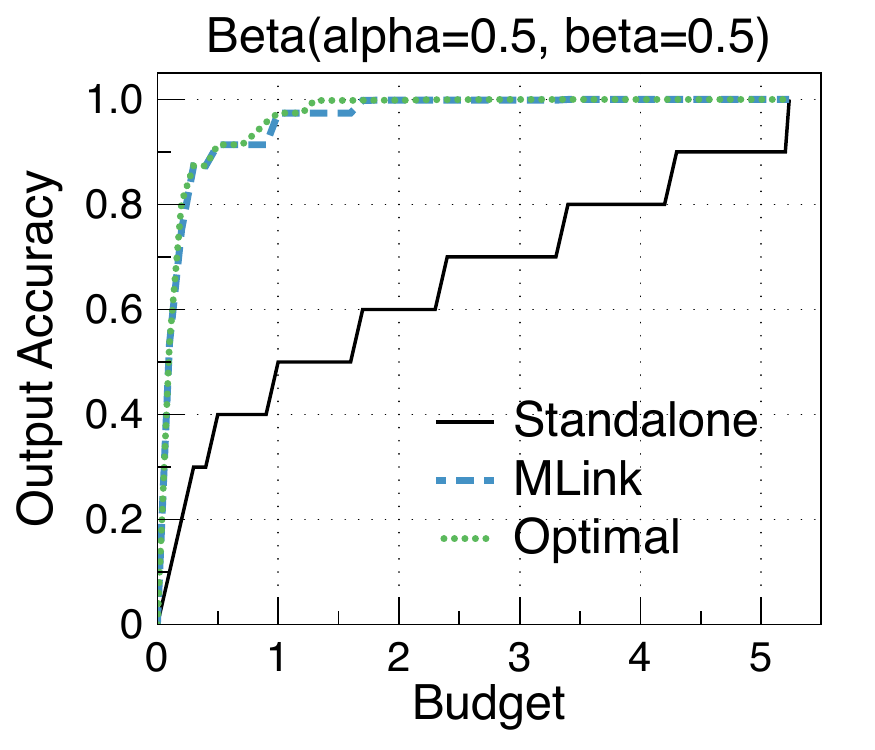}
         \caption{Beta Distribution}
         \label{fig:beta}
     \end{subfigure}
\caption{Simulation results of scheduling 10 models.}
\label{fig:simu}
\end{figure}

\textbf{Scalability.}
We did simulations to test the scheduling performance when the number of models is large.
Fig.~\ref{fig:simu} shows the comparisons of Standalone, \textit{MLink}, and optimal schedules on two simulated cases.
In order to simulate normal and relatively extreme conditions of model linking, we generated performances of model links among 10 models and their costs with the normal distribution (0.5 mean, 0.2 std) and beta distribution (0.5 alpha, 0.5 beta) , respectively.
And we set the ensemble gain fixed as 0.02.
Both the optimal and standalone schedules are found by brute-force enumeration.
Experimental results show that \textit{MLink} achieves near-optimal scheduling results and significantly outperforms the standalone baseline.

\begin{table}[t]
\centering
\caption{Comparisons of \textit{MLink}, MTL, Reducto, DRLS, and Standalone methods on two video analytics systems. 
}
\begin{tabular}{@{}l|ll|ll@{}}
\toprule
\multirow{2}{*}{Method} & \multicolumn{2}{c|}{Building (5/9GB Mem.)} & \multicolumn{2}{c}{City (5/9GB Mem.)} \\ \cmidrule(l){2-5} 
 & Acc. (\%) & Time (ms) & Acc. (\%) & Time (ms) \\ \midrule
Standalone & 33.3/66.7 & 30/74 & 33.3/66.7 & 55/121 \\ \midrule
MTL & 53.3 & 32.8 & 61.3 & 32.5 \\
DRLS & 45.7/81.3 & 58.7/107 & 39.5/77.6 & 102/188 \\
Reducto & 91.8/96.9 & 45.7/89 & 84.1/95.3 & 64/127 \\ \midrule
\textit{MLink} & \textbf{94.1/97.9} & 39.3/84 & \textbf{94/97.4} & 62/125 \\ \bottomrule
\end{tabular}
\label{tab:schedule}
\end{table}

\begin{figure}[t]
    \centering
    \includegraphics[width=0.8\linewidth]{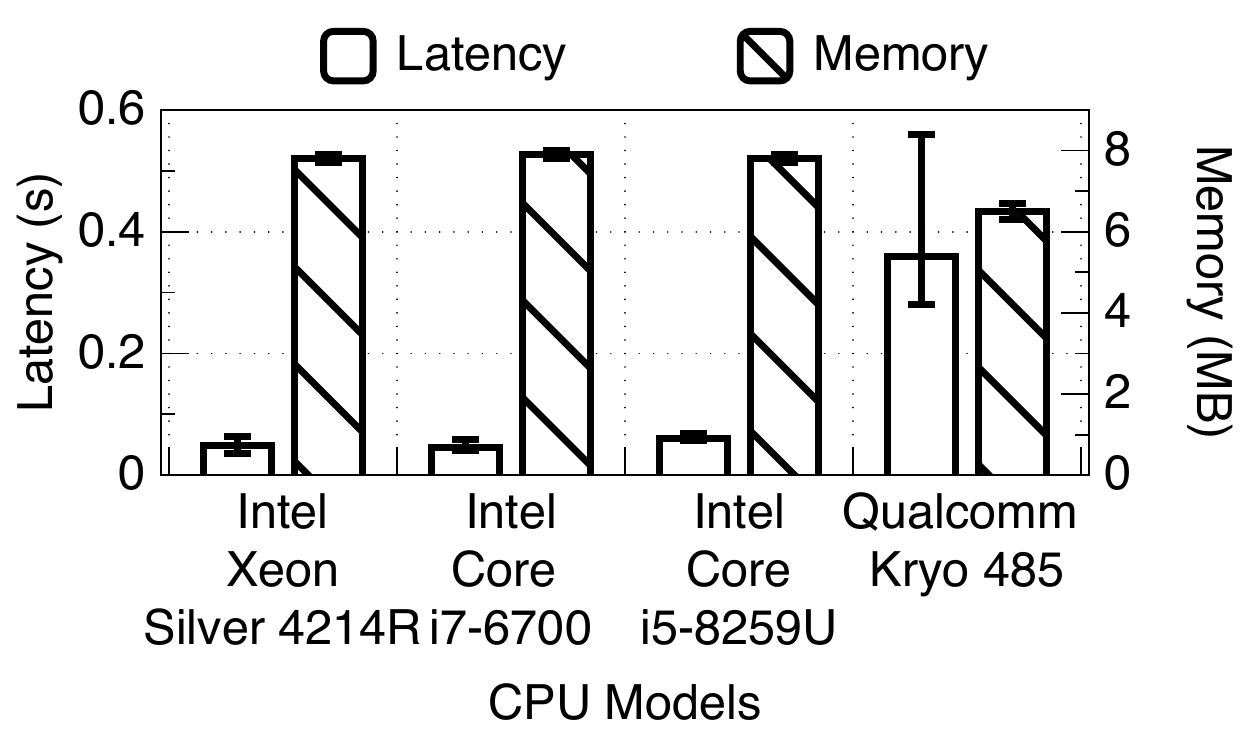}
\caption{\textit{MLink}'s overheads on servers and mobile phones.}
\label{fig:cpu}
\end{figure}

\textbf{Overheads of \textit{MLink}.}
In \textit{MLink}, model link's training and inference actions have good concurrency and efficiency.
We tested concurrent model link training and inference with input and output length randomly set from 1 to 100.
The average training time per process decreases to less than one second with more than 20 concurrent training processes.
And given 100 concurrent processes with one million samples to inference, the overall latency is only around one minute.
We deployed \textit{MLink} on four different devices (a cloud server, an edge server, a laptop, and a mobile phone) and tested its latency and memory footprint. 
As shown in Fig.~\ref{fig:cpu}, \textit{MLink} only introduces negligible additional overheads.
We also tested the communication overhead of aggregating model links of action classification models. 
For the 3-client case, the overall communication cost is 88*3 (server broadcast) + 88*3 (client update) = 528 bytes per sample. 
We set the batch size as 32, which translates into a communication cost of 16.5 KB per training step.

\section{Related Work}

\textbf{Multi-task learning and zipping.}
One straightforward way to optimize multiple standalone ML models is multi-task learning~\cite{hierarchical-mtl,mtl,mtl-survey} and zipping~\cite{multitask-zip}.
By sharing the same backbone neural networks among different tasks, multi-task models can provide richer inference results than standalone models under the same cost budget.
However, multi-task learning approaches lack flexibility and scalability, i.e., we need to tailor multi-task solutions case by case and re-design once the set of tasks change.
In contrast, although there exists parameter redundancy between different black-box ML models, \textit{MLink} approach can be flexibly extended.
And experiments show that the accuracy of model linking is higher when given a small amount of training data.

\textbf{Knowledge distillation.}
Following the taxonomy in the recent survey~\cite{kd-survey}, knowledge distillation has three main sources of knowledge: 
(1) response~\cite{response-1, kd}: the output of the ``teacher'' model;
(2) feature~\cite{feature-1}: the intermediate feature maps;
(3) relation~\cite{relation-1}: the relations of feature maps.
Mutual distillation~\cite{dml,mutual-distill} was proposed to train an ensemble of ``student'' models and let them learn from each other mutually.
Cross-task distillation~\cite{cross-task-distill} was proposed to train a ``student'' model by a ``teacher'' model pre-trained for another task.
Model linking is complementary to knowledge distillation, i.e. model links can be built among distilled ``student'' models.

\textbf{Redundancy filtering.}
Filtering redundant computation or communication is a promising way toward cost-efficient inference.
Yuan et al.~\cite{adams} proposed a reinforcement learning-based scheduler for multi-model data labeling tasks, which leverages the executed models' outputs as the hint information to schedule remaining models.
DNNs-aware video streaming~\cite{comm-srccompress} was proposed to compress the pixels less related with inference accuracy for communication-efficient inference.
FoggyCache~\cite{reuse-foggycache} reuses cached inference results by adaptive hashing input values.
Our proposed \textit{MLink} scheduler optimizes the cost-efficiency in a novel and more direct way: predict inference results of unexecuted ML models by executed models' outputs.

\section{Conclusion}
In this work, we propose to link black-box ML models and present the designs of model links and a collaborative multi-model inference algorithm.
The comprehensive evaluations show the effectiveness of black-box model linking and the superiority of the \textit{MLink} compared to other alternative methods.
We summarize limitations and future work as follows:
(1) When the semantic correlations between source and target models are low, model linking has poor output accuracy.
(2) When the number of joined models is very large, pairwise model linking will become unpractical.
So we will study how to smartly select models to build model links in the future.

\ifCLASSOPTIONcompsoc
  \section*{Acknowledgments}
\else
  \section*{Acknowledgment}
\fi

This research was supported by the National Key R\&D Program of China 2021YFB2900103, China National Natural Science Foundation with No. 61932016, No. 62132018.
This work is partially sponsored by CAAI-Huawei MindSpore Open Fund and ``the Fundamental Research Funds for the Central Universities'' WK2150110024.

\ifCLASSOPTIONcaptionsoff
  \newpage
\fi


\bibliographystyle{IEEEtran}
%

\bibliography{reference.bib}

\begin{thebibliography}{10}
\providecommand{\url}[1]{#1}
\csname url@samestyle\endcsname
\providecommand{\newblock}{\relax}
\providecommand{\bibinfo}[2]{#2}
\providecommand{\BIBentrySTDinterwordspacing}{\spaceskip=0pt\relax}
\providecommand{\BIBentryALTinterwordstretchfactor}{4}
\providecommand{\BIBentryALTinterwordspacing}{\spaceskip=\fontdimen2\font plus
\BIBentryALTinterwordstretchfactor\fontdimen3\font minus
  \fontdimen4\font\relax}
\providecommand{\BIBforeignlanguage}[2]{{%
\expandafter\ifx\csname l@#1\endcsname\relax
\typeout{** WARNING: IEEEtran.bst: No hyphenation pattern has been}%
\typeout{** loaded for the language `#1'. Using the pattern for}%
\typeout{** the default language instead.}%
\else
\language=\csname l@#1\endcsname
\fi
#2}}
\providecommand{\BIBdecl}{\relax}
\BIBdecl

\bibitem{mlink}
\BIBentryALTinterwordspacing
M.~Yuan, L.~Zhang, and X.-Y. Li, ``Mlink: Linking black-box models for
  collaborative multi-model inference,'' \emph{Proceedings of the AAAI
  Conference on Artificial Intelligence}, vol.~36, no.~9, pp. 9475--9483, Jun.
  2022. [Online]. Available:
  \url{https://ojs.aaai.org/index.php/AAAI/article/view/21180}
\BIBentrySTDinterwordspacing

\bibitem{smart-speaker}
F.~Bentley, C.~Luvogt, M.~Silverman, R.~Wirasinghe, B.~White, and D.~Lottridge,
  ``Understanding the long-term use of smart speaker assistants,''
  \emph{Proceedings of the ACM on Interactive, Mobile, Wearable and Ubiquitous
  Technologies}, vol.~2, no.~3, pp. 1--24, 2018.

\bibitem{smartcity}
L.~Duan, Y.~Lou, S.~Wang, W.~Gao, and Y.~Rui, ``Ai-oriented large-scale video
  management for smart city: Technologies, standards, and beyond,'' \emph{IEEE
  MultiMedia}, vol.~26, no.~2, pp. 8--20, 2018.

\bibitem{drone-video}
N.~Dilshad, J.~Hwang, J.~Song, and N.~Sung, ``Applications and challenges in
  video surveillance via drone: A brief survey,'' in \emph{2020 International
  Conference on Information and Communication Technology Convergence
  (ICTC)}.\hskip 1em plus 0.5em minus 0.4em\relax IEEE, 2020, pp. 728--732.

\bibitem{multi-modal-auto-drive}
D.~Feng, C.~Haase-Sch{\"u}tz, L.~Rosenbaum, H.~Hertlein, C.~Glaeser, F.~Timm,
  W.~Wiesbeck, and K.~Dietmayer, ``Deep multi-modal object detection and
  semantic segmentation for autonomous driving: Datasets, methods, and
  challenges,'' \emph{IEEE Transactions on Intelligent Transportation Systems},
  vol.~22, no.~3, pp. 1341--1360, 2020.

\bibitem{multitask-zip}
X.~He, Z.~Zhou, and L.~Thiele, ``Multi-task zipping via layer-wise neuron
  sharing,'' in \emph{Proceedings of the 32nd International Conference on
  Neural Information Processing Systems}, 2018, pp. 6019--6029.

\bibitem{hierarchical-mtl}
V.~Sanh, T.~Wolf, and S.~Ruder, ``A hierarchical multi-task approach for
  learning embeddings from semantic tasks,'' in \emph{Proceedings of the AAAI
  Conference on Artificial Intelligence}, 2019, pp. 6949--6956.

\bibitem{mtl-survey}
M.~Crawshaw, ``Multi-task learning with deep neural networks: A survey,''
  \emph{arXiv preprint arXiv:2009.09796}, 2020.

\bibitem{mtl}
Y.~Zhang and Q.~Yang, ``A survey on multi-task learning,'' \emph{IEEE
  Transactions on Knowledge and Data Engineering}, 2021.

\bibitem{kd}
G.~Hinton, O.~Vinyals, and J.~Dean, ``Distilling the knowledge in a neural
  network,'' \emph{arXiv preprint arXiv:1503.02531}, 2015.

\bibitem{resource-adadeep}
S.~Liu, Y.~Lin, Z.~Zhou, K.~Nan, H.~Liu, and J.~Du, ``On-demand deep model
  compression for mobile devices: A usage-driven model selection framework,''
  in \emph{Proceedings of the 16th Annual International Conference on Mobile
  Systems, Applications, and Services}, 2018, pp. 389--400.

\bibitem{adversarial-distillation}
M.~Goldblum, L.~Fowl, S.~Feizi, and T.~Goldstein, ``Adversarially robust
  distillation,'' in \emph{Proceedings of the AAAI Conference on Artificial
  Intelligence}, vol.~34, no.~04, 2020, pp. 3996--4003.

\bibitem{few-shot-compress}
H.~Bai, J.~Wu, I.~King, and M.~Lyu, ``Few shot network compression via cross
  distillation,'' in \emph{Proceedings of the AAAI Conference on Artificial
  Intelligence}, vol.~34, no.~04, 2020, pp. 3203--3210.

\bibitem{reuse-foggycache}
P.~Guo, B.~Hu, R.~Li, and W.~Hu, ``Foggycache: Cross-device approximate
  computation reuse,'' in \emph{Proceedings of the 24th Annual International
  Conference on Mobile Computing and Networking}, 2018, pp. 19--34.

\bibitem{reuse-deep}
L.~{Ning}, H.~{Guan}, and X.~{Shen}, ``Adaptive deep reuse: Accelerating cnn
  training on the fly,'' in \emph{2019 IEEE 35th International Conference on
  Data Engineering (ICDE)}, 2019, pp. 1538--1549.

\bibitem{reducto}
Y.~Li, A.~Padmanabhan, P.~Zhao, Y.~Wang, G.~H. Xu, and R.~Netravali, ``Reducto:
  On-camera filtering for resource-efficient real-time video analytics,'' in
  \emph{Proceedings of the Annual conference of the ACM Special Interest Group
  on Data Communication on the applications, technologies, architectures, and
  protocols for computer communication}, 2020, pp. 359--376.

\bibitem{adaconf}
J.~Jiang, G.~Ananthanarayanan, P.~Bodik, S.~Sen, and I.~Stoica, ``Chameleon:
  scalable adaptation of video analytics,'' in \emph{Proceedings of the 2018
  Conference of the ACM Special Interest Group on Data Communication}, 2018,
  pp. 253--266.

\bibitem{adams}
M.~Yuan, L.~Zhang, X.-Y. Li, and H.~Xiong, ``Comprehensive and efficient data
  labeling via adaptive model scheduling,'' in \emph{2020 IEEE 36th
  International Conference on Data Engineering (ICDE)}.\hskip 1em plus 0.5em
  minus 0.4em\relax IEEE, 2020, pp. 1858--1861.

\bibitem{Song2020Overlearning}
C.~Song and V.~Shmatikov, ``Overlearning reveals sensitive attributes,'' in
  \emph{International Conference on Learning Representations}, 2020.

\bibitem{tl-survey}
C.~Tan, F.~Sun, T.~Kong, W.~Zhang, C.~Yang, and C.~Liu, ``A survey on deep
  transfer learning,'' in \emph{International conference on artificial neural
  networks}.\hskip 1em plus 0.5em minus 0.4em\relax Springer, 2018, pp.
  270--279.

\bibitem{wang2018deep}
M.~Wang and W.~Deng, ``Deep visual domain adaptation: A survey,''
  \emph{Neurocomputing}, vol. 312, pp. 135--153, 2018.

\bibitem{online-kd}
R.~T. Mullapudi, S.~Chen, K.~Zhang, D.~Ramanan, and K.~Fatahalian, ``Online
  model distillation for efficient video inference,'' in \emph{Proceedings of
  the IEEE/CVF International Conference on Computer Vision}, 2019, pp.
  3573--3582.

\bibitem{video-ads}
M.~Yuan, L.~Zhang, Z.~Wu, and D.~Zheng, ``High-quality activity-level video
  advertising,'' in \emph{2020 IEEE/ACM 28th International Symposium on Quality
  of Service (IWQoS)}.\hskip 1em plus 0.5em minus 0.4em\relax IEEE, 2020, pp.
  1--10.

\bibitem{zero-shot-event}
M.~Elhoseiny, J.~Liu, H.~Cheng, H.~Sawhney, and A.~Elgammal, ``Zero-shot event
  detection by multimodal distributional semantic embedding of videos,'' in
  \emph{Thirtieth AAAI Conference on Artificial Intelligence}, 2016.

\bibitem{multimodal-survey}
T.~Baltru{\v{s}}aitis, C.~Ahuja, and L.-P. Morency, ``Multimodal machine
  learning: A survey and taxonomy,'' \emph{IEEE transactions on pattern
  analysis and machine intelligence}, vol.~41, no.~2, pp. 423--443, 2018.

\bibitem{black2002multi}
J.~Black, T.~Ellis, and P.~Rosin, ``Multi view image surveillance and
  tracking,'' in \emph{Workshop on Motion and Video Computing, 2002.
  Proceedings.}\hskip 1em plus 0.5em minus 0.4em\relax IEEE, 2002, pp.
  169--174.

\bibitem{wang2020alignnet}
J.~Wang, Z.~Fang, and H.~Zhao, ``Alignnet: A unifying approach to audio-visual
  alignment,'' in \emph{Proceedings of the IEEE/CVF Winter Conference on
  Applications of Computer Vision}, 2020, pp. 3309--3317.

\bibitem{moe}
S.~E. Yuksel, J.~N. Wilson, and P.~D. Gader, ``Twenty years of mixture of
  experts,'' \emph{IEEE transactions on neural networks and learning systems},
  vol.~23, no.~8, pp. 1177--1193, 2012.

\bibitem{hjelm2018learning}
\BIBentryALTinterwordspacing
R.~D. Hjelm, A.~Fedorov, S.~Lavoie-Marchildon, K.~Grewal, P.~Bachman,
  A.~Trischler, and Y.~Bengio, ``Learning deep representations by mutual
  information estimation and maximization,'' in \emph{International Conference
  on Learning Representations}, 2019. [Online]. Available:
  \url{https://openreview.net/forum?id=Bklr3j0cKX}
\BIBentrySTDinterwordspacing

\bibitem{gcam}
R.~R. Selvaraju, M.~Cogswell, A.~Das, R.~Vedantam, D.~Parikh, and D.~Batra,
  ``Grad-cam: Visual explanations from deep networks via gradient-based
  localization,'' \emph{International Journal of Computer Vision}, vol. 128,
  no.~2, pp. 336--359, 2020.

\bibitem{yolov3}
J.~Redmon and A.~Farhadi, ``Yolov3: An incremental improvement,'' \emph{arXiv
  preprint arXiv:1804.02767}, 2018.

\bibitem{resnet}
K.~He, X.~Zhang, S.~Ren, and J.~Sun, ``Deep residual learning for image
  recognition,'' in \emph{Proceedings of the IEEE conference on computer vision
  and pattern recognition}, 2016, pp. 770--778.

\bibitem{finetune}
Y.~Guo, H.~Shi, A.~Kumar, K.~Grauman, T.~Rosing, and R.~Feris, ``Spottune:
  transfer learning through adaptive fine-tuning,'' in \emph{Proceedings of the
  IEEE/CVF Conference on Computer Vision and Pattern Recognition}, 2019, pp.
  4805--4814.

\bibitem{tripuraneni2020theory}
N.~Tripuraneni, M.~Jordan, and C.~Jin, ``On the theory of transfer learning:
  The importance of task diversity,'' \emph{Advances in Neural Information
  Processing Systems}, vol.~33, 2020.

\bibitem{lstm}
S.~Hochreiter and J.~Schmidhuber, ``Long short-term memory,'' \emph{Neural
  computation}, vol.~9, no.~8, pp. 1735--1780, 1997.

\bibitem{attention}
D.~Bahdanau, K.~H. Cho, and Y.~Bengio, ``Neural machine translation by jointly
  learning to align and translate,'' in \emph{3rd International Conference on
  Learning Representations, ICLR 2015}, 2015.

\bibitem{seq2seq}
I.~Sutskever, O.~Vinyals, and Q.~V. Le, ``Sequence to sequence learning with
  neural networks,'' in \emph{Advances in neural information processing
  systems}, 2014, pp. 3104--3112.

\bibitem{multi-model-ensemble}
Z.~Shen, Z.~He, and X.~Xue, ``Meal: Multi-model ensemble via adversarial
  learning,'' in \emph{Proceedings of the AAAI Conference on Artificial
  Intelligence}, 2019, pp. 4886--4893.

\bibitem{li2021online}
Z.~Li, J.~Ye, M.~Song, Y.~Huang, and Z.~Pan, ``Online knowledge distillation
  for efficient pose estimation,'' in \emph{Proceedings of the IEEE/CVF
  International Conference on Computer Vision}, 2021, pp. 11\,740--11\,750.

\bibitem{active-learning-survey}
B.~Settles, ``Active learning literature survey,'' University of
  Wisconsin-Madison Department of Computer Sciences, Tech. Rep., 2009.

\bibitem{regression-active}
D.~J. MacKay, ``Information-based objective functions for active data
  selection,'' \emph{Neural computation}, vol.~4, no.~4, pp. 590--604, 1992.

\bibitem{loss-pred}
D.~Yoo and I.~S. Kweon, ``Learning loss for active learning,'' in
  \emph{Proceedings of the IEEE/CVF conference on computer vision and pattern
  recognition}, 2019, pp. 93--102.

\bibitem{rangwani2022closer}
H.~Rangwani, S.~K. Aithal, M.~Mishra, A.~Jain, and V.~B. Radhakrishnan, ``A
  closer look at smoothness in domain adversarial training,'' in
  \emph{International Conference on Machine Learning}.\hskip 1em plus 0.5em
  minus 0.4em\relax PMLR, 2022, pp. 18\,378--18\,399.

\bibitem{singh2021clda}
A.~Singh, ``Clda: Contrastive learning for semi-supervised domain adaptation,''
  \emph{Advances in Neural Information Processing Systems}, vol.~34, pp.
  5089--5101, 2021.

\bibitem{ng2021federated}
D.~Ng, X.~Lan, M.~M.-S. Yao, W.~P. Chan, and M.~Feng, ``Federated learning: a
  collaborative effort to achieve better medical imaging models for individual
  sites that have small labelled datasets,'' \emph{Quantitative Imaging in
  Medicine and Surgery}, vol.~11, no.~2, p. 852, 2021.

\bibitem{fl-concept}
Q.~Yang, Y.~Liu, T.~Chen, and Y.~Tong, ``Federated machine learning: Concept
  and applications,'' \emph{ACM Transactions on Intelligent Systems and
  Technology (TIST)}, vol.~10, no.~2, pp. 1--19, 2019.

\bibitem{FL-DA}
D.~Peterson, P.~Kanani, and V.~J. Marathe, ``Private federated learning with
  domain adaptation,'' \emph{arXiv preprint arXiv:1912.06733}, 2019.

\bibitem{zhou2012ensemble}
Z.-H. Zhou, \emph{Ensemble methods: foundations and algorithms}.\hskip 1em plus
  0.5em minus 0.4em\relax CRC press, 2012.

\bibitem{gender}
\BIBentryALTinterwordspacing
A.~Kumar, ``Pygender-voice,'' 2021, accessed: 2021-08-01. [Online]. Available:
  \url{https://github.com/abhijeet3922/PyGender-Voice}
\BIBentrySTDinterwordspacing

\bibitem{c3d}
D.~Tran, L.~Bourdev, R.~Fergus, L.~Torresani, and M.~Paluri, ``Learning
  spatiotemporal features with 3d convolutional networks,'' in
  \emph{Proceedings of the IEEE international conference on computer vision},
  2015, pp. 4489--4497.

\bibitem{face}
S.~I. Serengil and A.~Ozpinar, ``Lightface: A hybrid deep face recognition
  framework,'' in \emph{2020 Innovations in Intelligent Systems and
  Applications Conference (ASYU)}.\hskip 1em plus 0.5em minus 0.4em\relax IEEE,
  2020, pp. 23--27.

\bibitem{age}
G.~Levi and T.~Hassner, ``Age and gender classification using convolutional
  neural networks,'' in \emph{Proceedings of the IEEE conference on computer
  vision and pattern recognition workshops}, 2015, pp. 34--42.

\bibitem{caption}
\BIBentryALTinterwordspacing
G.~Wang, ``Image captioning,'' 2021, accessed: 2021-08-01. [Online]. Available:
  \url{https://github.com/DeepRNN/image\_captioning}
\BIBentrySTDinterwordspacing

\bibitem{speech}
\BIBentryALTinterwordspacing
Mozilla, ``Deepspeech,'' 2021, accessed: 2021-08-01. [Online]. Available:
  \url{https://github.com/mozilla/DeepSpeech}
\BIBentrySTDinterwordspacing

\bibitem{submodule-note}
M.~Sviridenko, ``A note on maximizing a submodular set function subject to a
  knapsack constraint,'' \emph{Operations Research Letters}, vol.~32, no.~1,
  pp. 41--43, 2004.

\bibitem{tf}
\BIBentryALTinterwordspacing
TensorFlow, ``Tensorflow,'' 2021, accessed: 2021-08-01. [Online]. Available:
  \url{https://github.com/tensorflow/tensorflow}
\BIBentrySTDinterwordspacing

\bibitem{pytorch}
\BIBentryALTinterwordspacing
PyTorch, ``Pytorch,'' 2021, accessed: 2021-08-01. [Online]. Available:
  \url{https://github.com/pytorch/pytorch}
\BIBentrySTDinterwordspacing

\bibitem{ms}
\BIBentryALTinterwordspacing
MindSpore, ``Mindsopre,'' 2021, accessed: 2021-08-01. [Online]. Available:
  \url{https://github.com/mindspore-ai/mindspore}
\BIBentrySTDinterwordspacing

\bibitem{hollywood2}
M.~Marszalek, I.~Laptev, and C.~Schmid, ``Actions in context,'' in \emph{2009
  IEEE Conference on Computer Vision and Pattern Recognition}.\hskip 1em plus
  0.5em minus 0.4em\relax IEEE, 2009, pp. 2929--2936.

\bibitem{openpose}
Z.~{Cao}, G.~{Hidalgo Martinez}, T.~{Simon}, S.~{Wei}, and Y.~A. {Sheikh},
  ``Openpose: Realtime multi-person 2d pose estimation using part affinity
  fields,'' \emph{IEEE Transactions on Pattern Analysis and Machine
  Intelligence}, 2019.

\bibitem{actionet}
\BIBentryALTinterwordspacing
M.~Olafenwa, ``Action-net,'' 2021, accessed: 2021-08-01. [Online]. Available:
  \url{https://github.com/OlafenwaMoses/Action-Net}
\BIBentrySTDinterwordspacing

\bibitem{traffic}
\BIBentryALTinterwordspacing
------, ``Traffic-net,'' 2021, accessed: 2021-08-01. [Online]. Available:
  \url{https://github.com/OlafenwaMoses/Traffic-Net}
\BIBentrySTDinterwordspacing

\bibitem{imagenet}
J.~Deng, W.~Dong, R.~Socher, L.-J. Li, K.~Li, and L.~Fei-Fei, ``Imagenet: A
  large-scale hierarchical image database,'' in \emph{2009 IEEE conference on
  computer vision and pattern recognition}.\hskip 1em plus 0.5em minus
  0.4em\relax Ieee, 2009, pp. 248--255.

\bibitem{rmsprop}
T.~Tieleman and G.~Hinton, ``Lecture 6.5-rmsprop: Divide the gradient by a
  running average of its recent magnitude,'' \emph{COURSERA: Neural networks
  for machine learning}, vol.~4, no.~2, pp. 26--31, 2012.

\bibitem{cityscapes}
M.~Cordts, M.~Omran, S.~Ramos, T.~Scharw{\"a}chter, M.~Enzweiler, R.~Benenson,
  U.~Franke, S.~Roth, and B.~Schiele, ``The cityscapes dataset,'' in \emph{CVPR
  Workshop on the Future of Datasets in Vision}, vol.~2.\hskip 1em plus 0.5em
  minus 0.4em\relax sn, 2015.

\bibitem{gtav}
S.~R. Richter, V.~Vineet, S.~Roth, and V.~Koltun, ``Playing for data: {G}round
  truth from computer games,'' in \emph{European Conference on Computer Vision
  (ECCV)}, ser. LNCS, B.~Leibe, J.~Matas, N.~Sebe, and M.~Welling, Eds., vol.
  9906.\hskip 1em plus 0.5em minus 0.4em\relax Springer International
  Publishing, 2016, pp. 102--118.

\bibitem{deeplabv3plus}
L.-C. Chen, Y.~Zhu, G.~Papandreou, F.~Schroff, and H.~Adam, ``Encoder-decoder
  with atrous separable convolution for semantic image segmentation,'' in
  \emph{Proceedings of the European conference on computer vision (ECCV)},
  2018, pp. 801--818.

\bibitem{office-home}
H.~Venkateswara, J.~Eusebio, S.~Chakraborty, and S.~Panchanathan, ``Deep
  hashing network for unsupervised domain adaptation,'' in \emph{({IEEE})
  Conference on Computer Vision and Pattern Recognition ({CVPR})}, 2017.

\bibitem{finetune-bestpractice}
B.~Chu, V.~Madhavan, O.~Beijbom, J.~Hoffman, and T.~Darrell, ``Best practices
  for fine-tuning visual classifiers to new domains,'' in \emph{European
  conference on computer vision}.\hskip 1em plus 0.5em minus 0.4em\relax
  Springer, 2016, pp. 435--442.

\bibitem{kd-survey}
J.~Gou, B.~Yu, S.~J. Maybank, and D.~Tao, ``Knowledge distillation: A survey,''
  \emph{International Journal of Computer Vision}, pp. 1--31, 2021.

\bibitem{response-1}
J.~Ba and R.~Caruana, ``Do deep nets really need to be deep?'' \emph{Advances
  in Neural Information Processing Systems}, vol.~27, 2014.

\bibitem{feature-1}
D.~Chen, J.-P. Mei, Y.~Zhang, C.~Wang, Z.~Wang, Y.~Feng, and C.~Chen,
  ``Cross-layer distillation with semantic calibration,'' in \emph{Proceedings
  of the AAAI Conference on Artificial Intelligence}, vol.~35, no.~8, 2021, pp.
  7028--7036.

\bibitem{relation-1}
N.~Passalis, M.~Tzelepi, and A.~Tefas, ``Heterogeneous knowledge distillation
  using information flow modeling,'' in \emph{Proceedings of the IEEE/CVF
  Conference on Computer Vision and Pattern Recognition}, 2020, pp. 2339--2348.

\bibitem{dml}
Y.~Zhang, T.~Xiang, T.~M. Hospedales, and H.~Lu, ``Deep mutual learning,'' in
  \emph{Proceedings of the IEEE Conference on Computer Vision and Pattern
  Recognition}, 2018, pp. 4320--4328.

\bibitem{mutual-distill}
A.~Yao and D.~Sun, ``Knowledge transfer via dense cross-layer
  mutual-distillation,'' in \emph{European Conference on Computer
  Vision}.\hskip 1em plus 0.5em minus 0.4em\relax Springer, 2020, pp. 294--311.

\bibitem{cross-task-distill}
H.-J. Ye, S.~Lu, and D.-C. Zhan, ``Distilling cross-task knowledge via
  relationship matching,'' in \emph{Proceedings of the IEEE/CVF Conference on
  Computer Vision and Pattern Recognition}, 2020, pp. 12\,396--12\,405.

\bibitem{comm-srccompress}
X.~Xie and K.-H. Kim, ``Source compression with bounded dnn perception loss for
  iot edge computer vision,'' in \emph{The 25th Annual International Conference
  on Mobile Computing and Networking}, 2019, pp. 1--16.

\end{thebibliography}

\newpage

\begin{IEEEbiography}[{
    \includegraphics[width=1in,height=1.25in,clip,keepaspectratio]{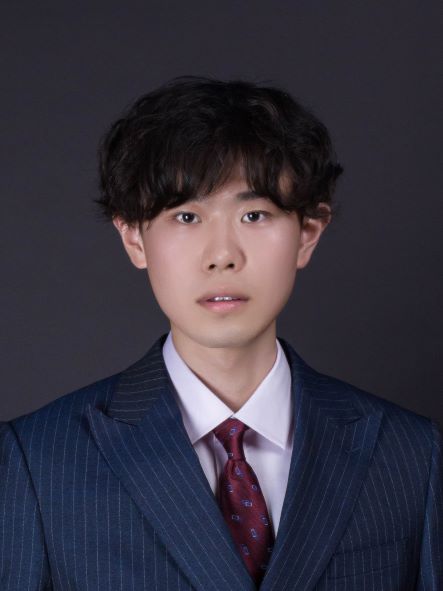}
}]{Mu Yuan}
is a Ph.D. candidate at the School of Computer Science and Technology, University of Science and Technology of China (USTC).
He received a bachelor's degree in computer science and technology from USTC.
His research interests include model inference and network systems.
\end{IEEEbiography}

\begin{IEEEbiography}[{
    \includegraphics[width=1in,height=1.25in,clip,keepaspectratio]{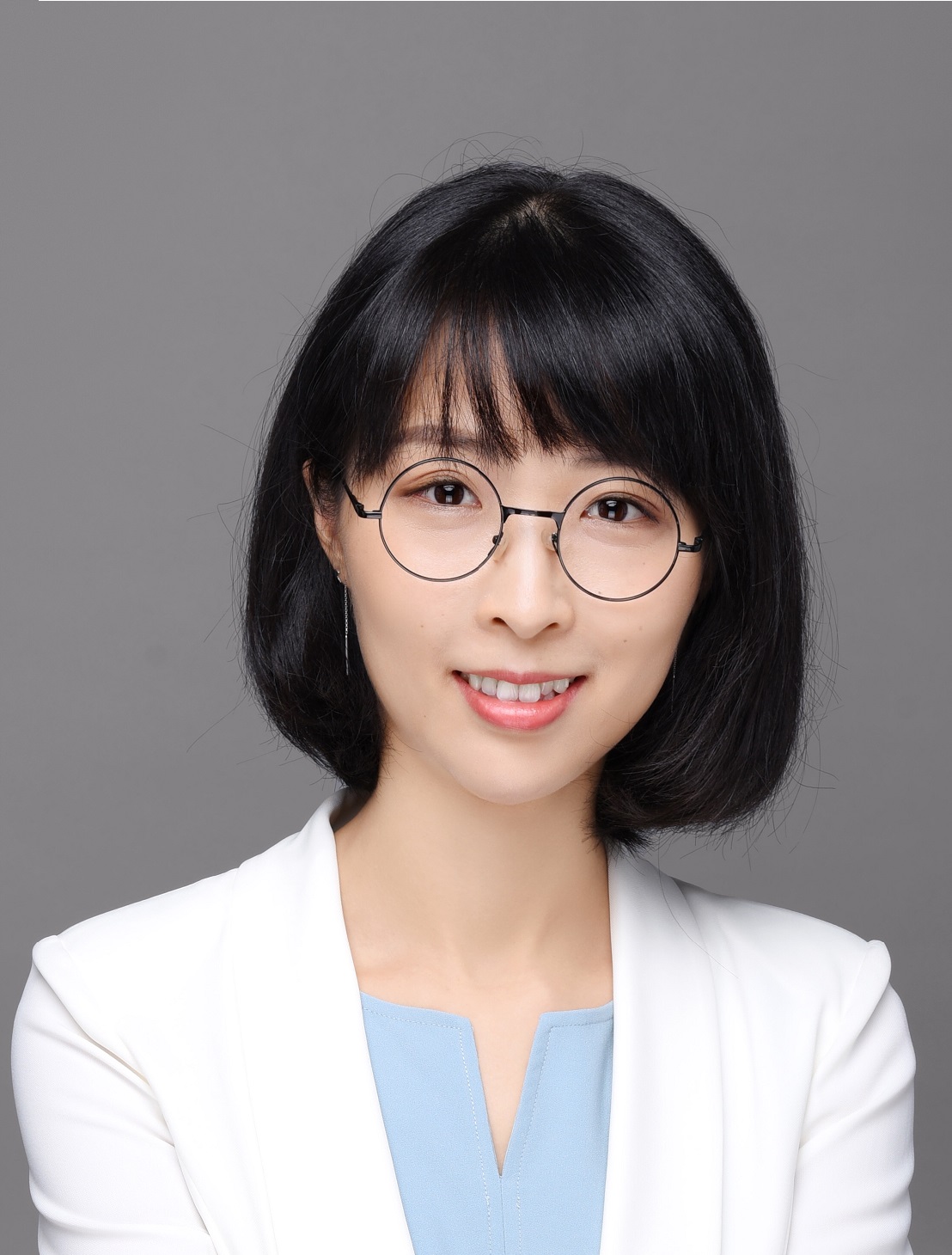}
}]{Lan Zhang}
is currently a Professor at the School of Computer Science and Technology, University of Science and Technology of China. She received her Ph.D degree and Bachelor degree from Tsinghua University, China. Her research interests include mobile computing, privacy protection,and data sharing and trading.
\end{IEEEbiography}

\begin{IEEEbiography}[{
    \includegraphics[width=1in,height=1.25in,clip,keepaspectratio]{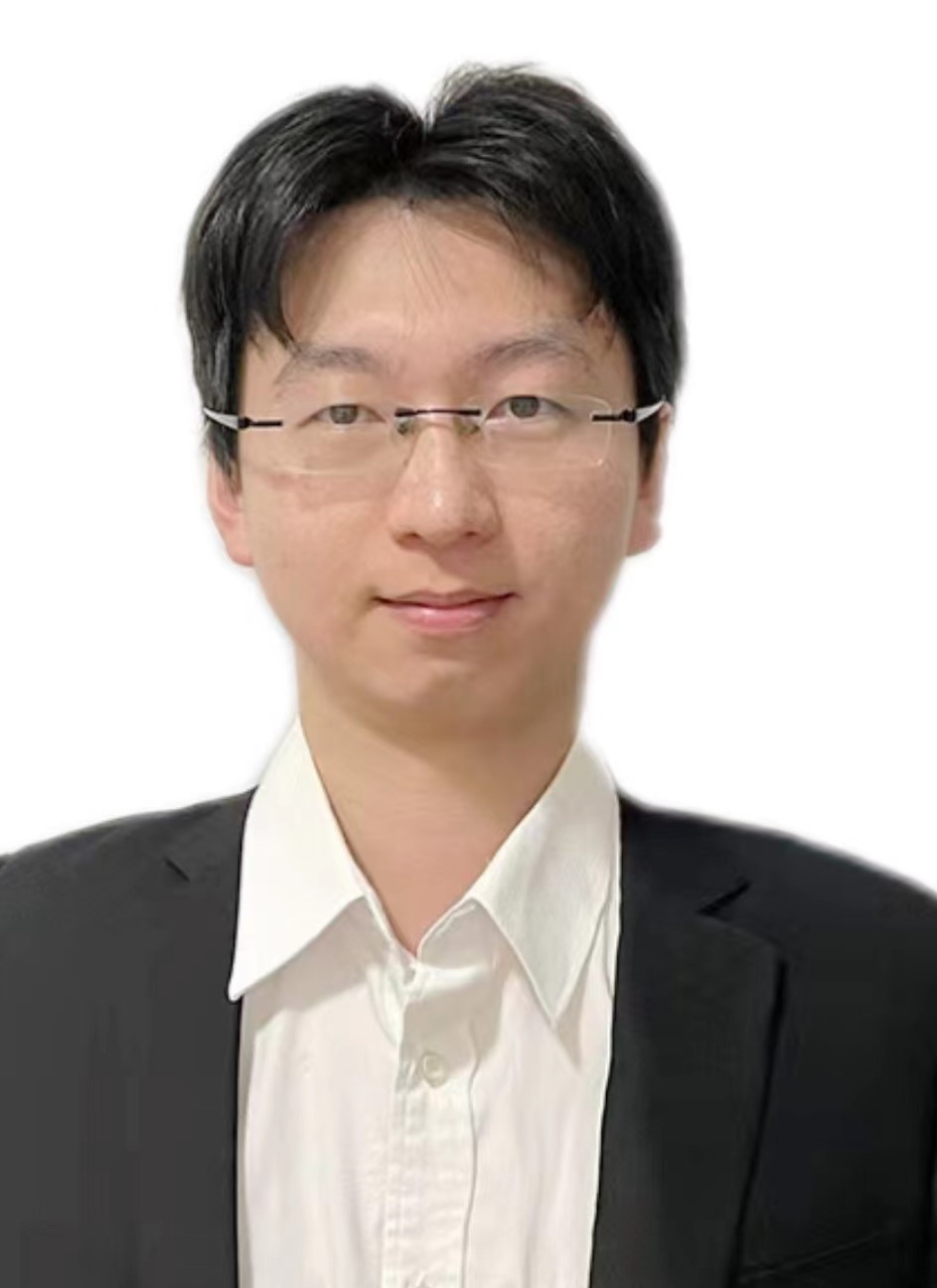}
}]{Zimu Zheng}
is currently a head research engineer at Huawei Cloud. He received his B.Eng. degree in South China University of Technology and Ph.D degree in the Hong Kong Polytechnic University. Zimu has received several awards for outstanding technical contributions in Huawei. He also received the Best Paper Award of ACM e-Energy and the Best Paper Award of ACM BuildSys in 2018. His research interest lies in edge intelligence, multi-task learning, and AIoT.
\end{IEEEbiography}

\begin{IEEEbiography}[{
    \includegraphics[width=1in,height=1.25in,clip,keepaspectratio]{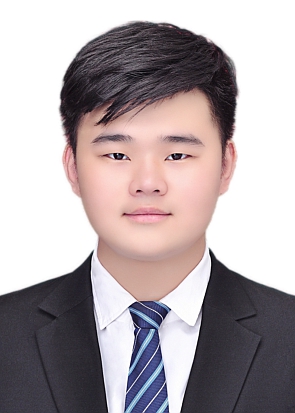}
}]{Yi-Nan Zhang}
is pursuing master degree at the School of Computer Science and Technology of China. He received his Bachelor degree from North Eastern University, China. His research interests include knowledge reasoning and multi models inference.
\end{IEEEbiography}

\begin{IEEEbiography}[{\includegraphics[width=1.0in,height=2.15in,clip,keepaspectratio]{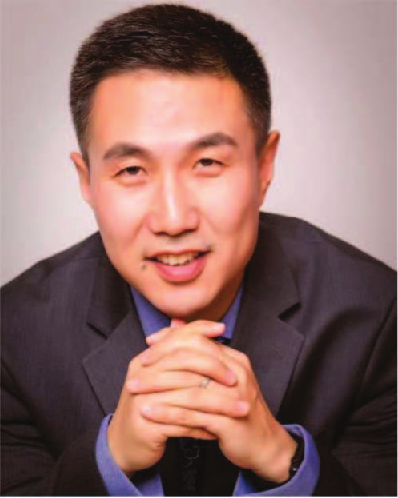}}]{Xiang-Yang Li}
 (Fellow, IEEE) is a professor and Executive Dean at School of Computer Science and Technology, USTC. He is an ACM Fellow (2019), IEEE fellow (2015), an ACM Distinguished Scientist (2014). He was a full professor at Computer Science Department of IIT and co-Chair of ACM China Council. Dr. Li received M.S. (2000) and Ph.D. (2001) degree at Department of Computer Science from University of Illinois at Urbana-Champaign. He received a Bachelor degree at Department of Computer Science from Tsinghua University, P.R. China, in 1995. His research interests include Artificial Intelligence of Things(AIOT), privacy and security of AIOT, and data sharing and trading.
\end{IEEEbiography}




\end{document}